\documentclass [preprint,12pt,hyperref,sort&compress] {elsarticle}

\usepackage{amssymb}
\usepackage{amsthm,amsmath,amssymb}
\usepackage{mathrsfs}
\usepackage{algorithmic}
\usepackage{algorithm}
\usepackage{graphicx}
\usepackage{enumitem}
\usepackage{booktabs} 
\usepackage{bm}
\usepackage{multirow}
\usepackage{makecell}
\usepackage{float}
\usepackage{booktabs}
\usepackage[font=small,labelfont=bf,labelsep=none]{caption}
\usepackage{booktabs}
\usepackage{graphicx}
\usepackage[
pdfauthor={derajan},
pdftitle={How to do this},
pdfstartview=XYZ,
bookmarks=true,
colorlinks=true,
linkcolor=blue,
urlcolor=blue,
citecolor=blue,
pdftex,
bookmarks=true,
linktocpage=true,   
hyperindex=true
]{hyperref}

\begin{document}

\begin{frontmatter}



\title{Trajectory Planning for Teleoperated Space Manipulators Using Deep Reinforcement Learning}


\affiliation[inst1]{organization={Shenzhen International Graduate
School, Tsinghua University},
            addressline={xiab21@mails.tsinghua.edu.cn}, 
            city={Shenzhen},
            postcode={518055}, 
            country={China}}
\affiliation[inst2]{organization={Shenzhen International Graduate
School, Tsinghua University},
            addressline={txr2023214329@gmail.com}, 
            city={Shenzhen},
            postcode={518055}, 
            country={China}}
\affiliation[inst3]{organization={Research Institute of Tsinghua University in Shenzhen},
            addressline={boyuan@ieee.org}, 
            city={Shenzhen},
            postcode={518057}, 
            country={China}}
\affiliation[inst4]{organization={Shenzhen International Graduate
School, Tsinghua University},
            addressline={zhhli@tsinghua.edu.cn}, 
            city={Shenzhen},
            postcode={518055}, 
            country={China}}
\affiliation[inst5]{organization={Department of Automation, Tsinghua University},
            addressline={liangbin@tsinghua.edu.cn}, 
            city={Beijing},
            postcode={100000}, 
            country={China}}
\affiliation[inst6]{organization={Shenzhen International Graduate
School, Tsinghua University},
            addressline={wang.xq@sz.tsinghua.edu.cn}, 
            city={Shenzhen},
            postcode={518055}, 
            country={China}}

\author[inst1]{Bo Xia} 
\author[inst2]{Xianru Tian}
\author[inst3]{Bo Yuan}
\author[inst4]{Zhiheng Li}
\author[inst5]{Bin Liang}
\author[inst6]{Xueqian Wang}

\begin{abstract}
Trajectory planning for teleoperated space manipulators involves challenges such as accurately modeling system dynamics, particularly in free-floating modes with non-holonomic constraints, and managing time delays that increase model uncertainty and affect control precision. 
Traditional teleoperation methods rely on precise dynamic models requiring complex parameter identification and calibration, while data-driven methods do not require prior knowledge but struggle with time delays. 
A novel framework utilizing deep reinforcement learning (DRL) is introduced to address these challenges. 
The framework incorporates three methods: Mapping, Prediction, and State Augmentation, to handle delays when delayed state information is received at the master end. 
The Soft Actor Critic (SAC) algorithm processes the state information to compute the next action, which is then sent to the remote manipulator for environmental interaction. 
Four environments are constructed using the MuJoCo simulation platform to account for variations in base and target fixation: fixed base and target, fixed base with rotated target, free-floating base with fixed target, and free-floating base with rotated target. 
Extensive experiments with both constant and random delays are conducted to evaluate the proposed methods. 
Results demonstrate that all three methods effectively address trajectory planning challenges, with State Augmentation showing superior efficiency and robustness.
\end{abstract}




\end{frontmatter}



\section{Introduction}
\label{Introuction}

With the aid of teleoperation technology, space robots significantly enhance astronauts' capabilities in space operations, playing an increasingly vital role in On-Orbit Servicing (OOS) missions, including tasks such as capturing, refueling, repairing satellites, removing orbital debris, and assembling and maintaining large space infrastructure \cite{gao2023hand, pryor2020interactive, zhang2022review, doggett2002robotic}. 
The current commonly used teleoperation control methods include teleprogramming control, bilateral control, and virtual predictive control\cite{wang2008general}.
Teleprogramming control operates in a supervisory mode, where space robots receive operational instructions from the master end and interact with the environment at the slave end, forming a closed-loop system\cite{zhang2017autonomous, funda1992teleprogramming}. 
However, this method relies on the intelligence level of the space robot.
Both bilateral control and virtual predictive control fall under the category of direct control. 
The distinction lies in bilateral control directly receives force feedback information from the remote environment and is suitable for scenarios with small delays. 
With the aid of appropriate control algorithms, such as passive control\cite{nuno2011passivity}, robust control\cite{wang2018novel, zhai2016adaptive, chen2014adaptive} and impendence control\cite{chen2016self, mersha2013bilateral, sharifi2018impedance}, the force and position information between the operator at the master end and the robot at the slave end is kept consistent. 
By contrast, virtual predictive control\cite{kheddar2007enhanced, deng2003predictive} establishes a virtual model at the master end similar to the environment and robot at the slave end, mitigating the impact of large delays on system stability and operational characteristics.
All model-based methods demand high model accuracy, while designing complex controllers entails deep domain expertise. 
However, space robots are intricate dynamic systems, presenting significant modeling challenges due to the complex coupling of dynamics between the base and the manipulator arm, as well as the nonlinearities stemming from factors such as friction and joint flexibility\cite{liu2015dynamics}. 
Furthermore, even with a physics-based model, many crucial parameters often remain unknown. These factors can have a substantial impact on the model and subsequently affect the performance of the controller.

Recently, data-driven model-free deep reinforcement learning (DRL) has demonstrated significant promise in various domains such as games\cite{mnih2013playing, vinyals2017starcraft}, industrial control\cite{park2022control, nian2020review}, and large language models\cite{ouyang2022training}. 
This approach has also been widely employed by scholars in the field of space robotics, predominantly focusing on trajectory planning for robotic arms.
For stationary targets, Yan et al.\cite{yan2018control} achieved single/dual-arm grasping using Soft Q learning. 
Wang et al.\cite{wang2021end} employed image inputs and the Soft Actor-Critic algorithm to decide joint angular velocities for controlling UR5 robotic arms. Lei et al.\cite{lei2022active} employed the Proximal Policy Optimization algorithm to achieve active object tracking with dynamically updated rewards to ensure that the target remains as centered as possible in the camera's field of view. Wu et al.\cite{wu2020reinforcement} conducted trajectory tasks effectively with slight target movement utilizing the Deep Deterministic Policy Gradient algorithm on a space robot with freely-floating dual arms.
To enhance the efficiency of DRL exploration and achieve more precise and stable results, Wang et al.\cite{wang2021multi} proposed an improved Proximal Policy Optimization algorithm, integrating methods for decision-making based on input states to obtain the final action. 
Cao et al.\cite{cao2023reinforcement} introduced inverse kinematics of fixed-base robots as a \textit{priori} strategies, guiding the intelligent agent toward the optimal strategy through a hybrid strategy construction process.
To avoid the potential collisions among robot arms, bases, and obstacles, Li et al.\cite{li2021constrained} modified the reward function based on the shortest distances between robot arm links and restrictions on end-effector velocity. 
Wang et al.\cite{wang2022collision} divided the entire system into two layers: the high-level policy for collision-free trajectory planning of end-effectors, and the low-level policy for decomposing any given pose into position and orientation sub-tasks.
Moreover, Wang et al.\cite{wang2022learning} addressed the issue of unknown motion states of non-cooperative target grasping points by predicting grasping points based on the object's point cloud information. 
Jiang et al.\cite{jiang2021coordinated} and Yang et al.\cite{yang2019control} further investigated the end-to-end control of flexible arms on space robots.
While previous studies have demonstrated the potential of DRL for space robotics, they typically assume a high level of autonomy on the remote robot, which is unrealistic given the current state of technology. In practice, the majority of intelligence resides on the master end, which introduces significant challenges for RL-based control due to communication delays and limited bandwidth.

A number of approaches have been proposed to address the issue of delay in reinforcement learning, which can be summarized into three main categories: information augmentation, model prediction, and others. 
Information augmentation methods primarily involve transforming the original delayed Markov decision process into a new delay-free Markov decision process based on an \textbf{Information State} composed of the most recent observed delay states and action sequences. 
The feasibility of this algorithm was initially analyzed theoretically by Katsikopoulos et al.\cite{katsikopoulos2003markov}, while Nath et al.\cite{nath2021revisiting} provided empirical evidences. 
Bouteiller et al.\cite{bouteiller2021reinforcement} performed partial trajectory resampling in delayed environments to effectively convert offline sub-trajectories into current policy-based sub-trajectories. 
Furthermore, leveraging information from agents in delay-free environments, such as a small amount of expert trajectories or previously learned expert policies, imitation learning can be employed to tackle tasks with delays\cite{xie2023addressing, liotet2022delayed}. 
Model prediction methods generally involve two steps: predicting unseen states due to delays and then making final decisions based on the predicted states and standard reinforcement learning algorithms. 
Therefore, simulating environmental dynamics accurately is crucial for this category of methods. 
Initially dominated by traditional algorithms, Walsh et al.\cite{walsh2009learning} viewed this process as a deterministic mapping, while Hester et al.\cite{hester2013texplore} used random forests to predict states and rewards. 
Subsequent developments based on deep neural networks saw Firoiu et al.\cite{firoiu2018human}, Derman et al.\cite{derman2020acting}, and Chen et al.\cite{chen2021delay} utilize recurrent neural networks, feed-forward models, and particle integration methods, respectively, to learn transitions. 
Additionally, using trajectories in delay-free environments, Xia et al. \cite{xia2024deer} learned a context representation encoder containing delay information, which was then used for encoding in delayed environments. 
Alternative approaches encompass two distinct categories: 
memoryless methods\cite{schuitema2010control}, which make decisions solely based on the latest observed state, ignoring any delay effects, and methods\cite{agarwal2021blind} that learn transition functions and Q-values in delay-free environments and select actions that may result in the maximum Q-values under predicted states during the decision-making process.

Motivated by the aforementioned methods, this study explores, for the first time, the application of DRL in trajectory planning for the remotely operated space manipulator.
The control process consists of three key components: a delay information processing module, a DRL decision module, and a remote environment interaction module.
The aim of the delay information processing module is to generate a state that facilitates current decision-making by considering both the current delayed state and historical action sequences.
To achieve this, three methods, Mapping, Prediction, and State Augmentation, are designed accordingly.
Following the completion of delay information processing, the decision module generates corresponding actions based on the obtained new state.
SAC is chosen as the decision-making algorithm due to its robustness in policy generation through cumulative rewards and maximum entropy optimization.
The remote environment interaction module operates in the MuJoCo simulation environment, including four single arm scenarios: fixed base and fixed target, fixed base and rotated target, free-floating base and fixed target, and free-floating base and rotated target.
Upon receiving the torque information from the master end, the space manipulator at the slave end interacts with the environment, generating the next state and reward, which are then fed back to the master end.

The paper is organized into five sections. 
Section \ref{sec:problem} introduces the research problem, including background knowledge such as the kinematics of space robot, the reinforcement learning setup, and models for both constant and random time delays in teleoperation.
Section \ref{sec:method} provides a detailed description of the entire control process, focusing particularly emphasis on the algorithmic design of the delay information processing module and the DRL decision module. 
Section \ref{sec:experiment} presents the simulation results under various delay scenarios across four distinct settings, along with an analysis of the outcomes. 
This paper is concluded in  Section \ref{sec:conclusion} with directions for future work.

\section{Problem Statement}
\label{sec:problem}
\subsection{Kinematics of space robot}
\begin{figure}[t]
    \centering
    \includegraphics[width=0.5\textwidth]{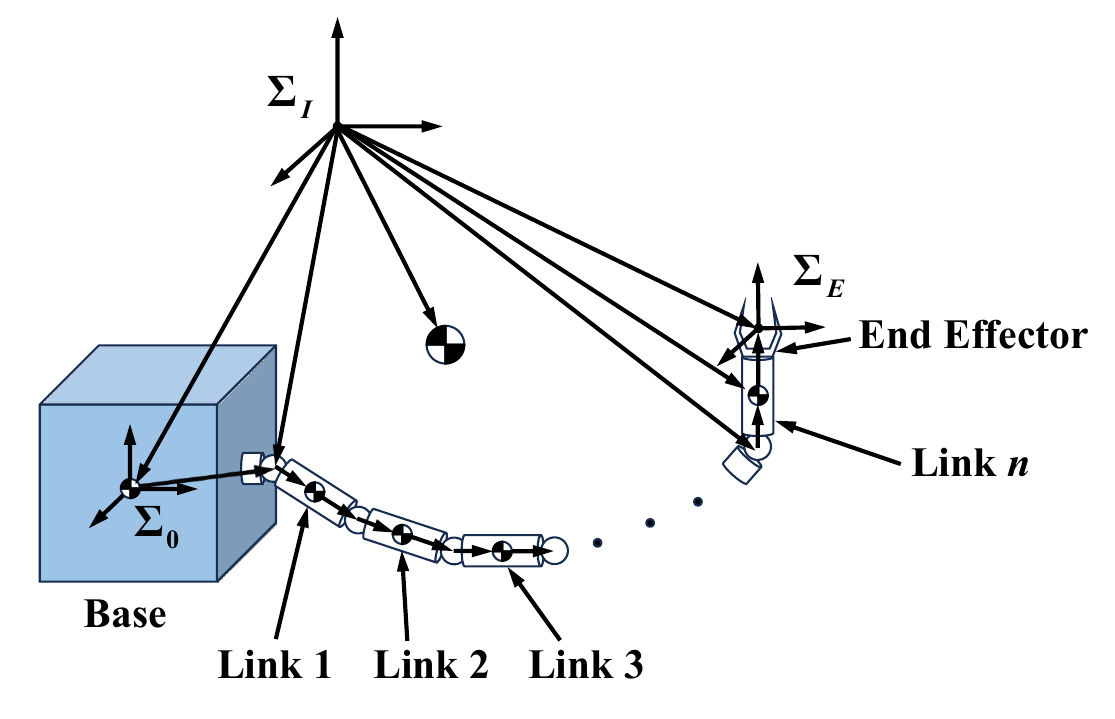}
    \caption{\ Schematic diagram of a space robot.}
    \label{Fig 1:space robot}
\end{figure}

According to Figure \ref{Fig 1:space robot}, a space robotic system typically consists of a spacecraft (base) and a manipulator with $n$ degrees of freedom (DOF). 

The vector of joint angles of the manipulator is denoted as $q = [\theta_1, \theta_2, \cdots, \theta_n] ^ {T}$, and correspondingly, its velocity can be represented as $\dot{q}$. Additionally, $(v_i, w_i)^{T}$ ($i=b, e$) represent the velocities of the base and the robotic arm, respectively. Referring to \cite{umetani1989resolved} , the following equation can be obtained:
\begin{equation}
\left (\begin{array}{c}
v_e \\
\omega_e  \\
\end{array}\right) = J_b \left (\begin{array}{c}
v_b  \\
\omega_b  \\
\end{array}\right)  + J_m \dot{q}\label{eq1},
\end{equation}
where $J_b$ and $J_m$ are the Jacobian matrices of the base and the manipulator respectively.
Without any external force or torques executed, the space robot is in free-floating mode and its linear momentum $P$ and angular momentum $L$ are conserved. 
Therefore, it holds that:
\begin{equation}
\left (\begin{array}{c}
P \\
L  \\
\end{array}\right) = H_b \left (\begin{array}{c}
v_b  \\
\omega_b  \\
\end{array}\right)  + H_{bm} \dot{q} \label{eq2},
\end{equation}
where $H_b$ and $H_{bm}$ are inertia matrix and coupling inertia matrix, 
respectively. 
Further, assuming the initial values of $P$ and $L$ are both zero, the velocity of the base can be expressed as:
\begin{equation}
\left (\begin{array}{c}
v_b \\
\omega_b   \\
\end{array}\right) = - H_b^{-1} H_{bm} \dot{q} = J_{bm} \dot{q}\label{eq3},
\end{equation}
where $J_{bm}$ is the Jacobian matrix of the base.
Substituting Eq. (\ref{eq3}) into Eq. (\ref{eq1}) yields the following expression:
\begin{equation}
\left (\begin{array}{c}
v_e \\
\omega_e  \\
\end{array}\right) = (J_b J_{bm} + J_m) \dot{q}=J_{g}\dot{q} \label{eq4},
\end{equation}
where $J_g$ is referred to as the generalized Jacobian matrix, which is not only related to kinematic parameters but also to dynamic parameters.

\subsection{Reinforcement learning}
\label{RL}
Reinforcement learning is a sequential decision-making process grounded in the theoretical framework of Markov decision processes (MDPs). Typically, an MDP is defined by a 6-element tuple $(\mathscr{S},\mathscr{A}, \mathscr{P}, \mathcal{R}, \rho, \gamma)$, where $\mathscr{S}$ and $\mathscr{A}$ represent the state space and action space, respectively; $\mathscr{P}$ is the state transition function and $\mathcal{R}$ denotes the reward function while $\rho$ represents the initial distribution of states; $\gamma$ signifies the discount factor for future rewards.
At time step $t$, the agent observes the environmental state $s_t$. Subsequently, based on its current policy $\pi$, it selects an action $a_t \sim \pi(\cdot | s_t)$ to apply to the environment. The environment then returns a new state $s_{t+1} \sim \mathscr{P}(\cdot | s_t, a_t)$ along with a reward $r_t = R(s_t, a_t)$. 
The objective of reinforcement learning is to maximize the cumulative reward $G_t = \sum_{k=t}^{T} \gamma^{k-t} r_k$ to obtain an optimal policy $\pi^*(\cdot | s_t)$, where $T$ is the maximum number of steps the agent interacts with the environment in an episode.

The trajectory planning problem of space robots can be modeled using MDPs, described by a tuple of six elements. Specifically, the setup for the state space, action space, and reward function is as follows\label{traditional MDP}:
\begin{itemize}[leftmargin=*]
    \item State $\mathscr{S}$. 
    
    Due to the crucial value of the information contained in the state for agent decision-making, it is imperative to consider as many relevant features influencing decisions as possible when designing the state. 
    In the trajectory planning process of space robots, factors such as the joint angles $q \in \mathbb{R}^n$, joint angular velocities
    $\dot{q} \in \mathbb{R}^n$, end-effector pose  $p_e \in \mathbb{R}^6$ and velocity $v_e \in \mathbb{R}^6$, target pose $p_{target} \in \mathbb{R}^6$, and the distance $d \in \mathbb{R}$ between the end-effector and the target point all play pivotal roles. 
    Particularly, when the space robot operates in a free-floating mode, the presence of non-holonomic constraints between the base and the manipulator can affect the manipulator's planning. Therefore, the base pose $p_b \in \mathbb{R}^6$ and velocity $v_b \in \mathbb{R}^6$ should also be included in the state space design. Consequently, for the fixed-base mode, the state is designed as follows:
    \begin{equation}
         s_t = (q, \dot{q} ,  p_e ,  v_e ,  p_{target},  d) \in \mathbb{R}^{2n+ 19};
    \end{equation}
    By contrast, for the free-floating mode, the state is designed as follows: 
    \begin{equation}
         s_t = (q, \dot{q} ,  p_e ,  v_e ,  p_b ,  v_b , p_{target},  d) \in \mathbb{R}^{2n+ 31}.
    \end{equation}
    \item Action $\mathscr{A}$. 
    
    The action is designed as a set of torques applied to the manipulator joints, clipped to a certain range: $a_t \in \mathbb{R}^n $ and  $a_{ti} \in [0, max(torque_i)](i=1, \cdots, n)$.
    \item Reward function $\mathcal{R}$. 
    
    The reward function is defined as follows:
     \begin{equation}
     \begin{aligned}
         r_t = & \omega_1 d_t + \omega_2 log(d_t + 1e^{-6}) + \omega_3 (||(v_e)_t|| + ||(v_b)_t||) + \omega_4 ||\dot{q_t}|| 
         \\ & +  \delta(d_t \le d_{threshold}) \omega_5 (T - t) log(d_{threshold}), \label{eq7}
     \end{aligned}
    \end{equation}
    where the episode is terminated when the distance $d$ is small than the threshold $d_{threshold}$. 
    Each term in Eq. \ref{eq7} serves a distinct purpose. The first term incentivizes the robotic arm's end effector to approach the target point as closely as possible. 
    The second term addresses the scenario where the end effector lingers near the target point as the distance nears the threshold without progressing further, ensuring continual advancement towards the target. 
    The third and fourth terms aim to mitigate excessive velocity fluctuations in the base, end-effector, and joints of the robotic arm during trajectory planning, promoting smooth transitions across these variables from a reward standpoint.
    Lastly, the fifth term functions as a terminal reward, activated only when the distance falls below or equals the threshold. 
    The magnitude of this reward is directly proportional to the remaining steps at the end of an episode,  $T - t$. 
    As a result, it is expected that as the agent completes tasks more swiftly, the reward value associated with this term will increase proportionally.
    
\end{itemize}

\begin{figure}[t]
    \centering
    \includegraphics[width=0.9\textwidth]{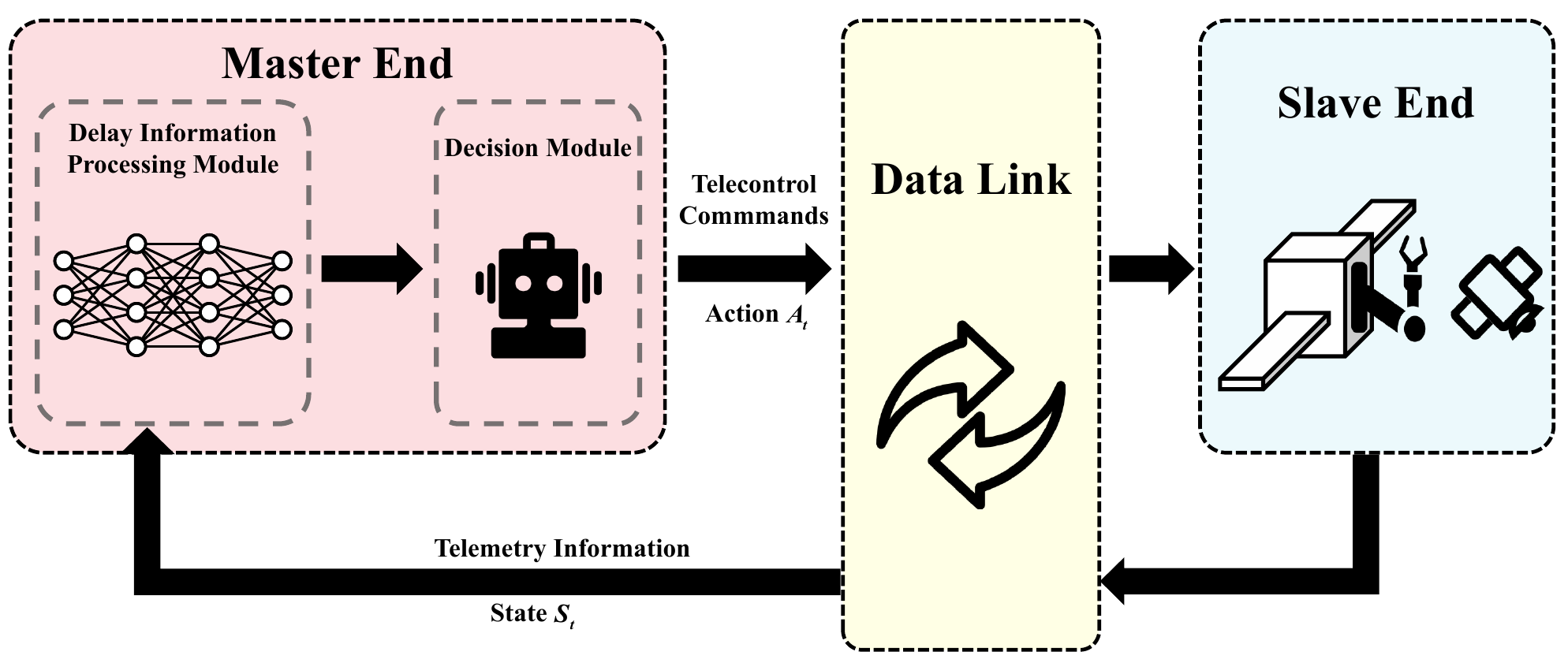}
    \caption{\ Overall Diagram of Space Teleoperation Based on DRL.}
    \label{Fig 2:spaceOverall Architecture}
\end{figure}

\subsection{Time delay in teleoperation}
In the context of space teleoperation, the time delay introduced by the data link disrupts the Markov property of the original decision-making process, significantly compromising the performance of reinforcement learning algorithms. 
Specifically, shown in Figure \ref{Fig 2:spaceOverall Architecture}, once an operator issues an instruction at the master end, there is a delay before this instruction reaches the slave end, which constitutes the action delay as referenced in \cite{katsikopoulos2003markov}. 
Meanwhile, when the space robot at the slave end actually interacts with the environment, the feedback from the environment, containing new states and rewards, is not promptly relayed to the master end, resulting in the observation delay and reward delay discussed in \cite{katsikopoulos2003markov}. 
Since observation delay and action delay exert equivalent influences on the agent's decision-making process, proven in \cite{katsikopoulos2003markov}, 
we solely focus on the observation delay, ensuring that its length aligns with that of the round time delay(RTD).

\textbf{Remark}.
At time $t$, the state generated by the slave end is denoted as $s_t$, while the state actually observed by the master end is denoted by $x_t$. 

\begin{figure}[t]
    \centering
    \includegraphics[width=0.6\textwidth]{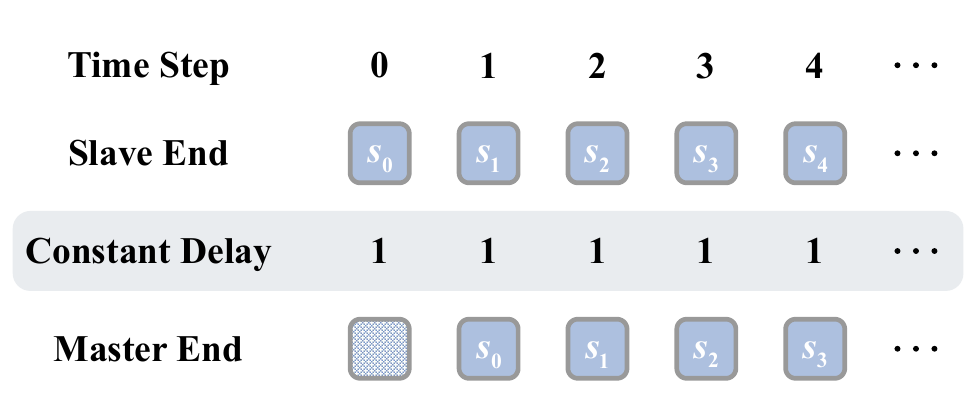}
    \caption{\ A constant delay scenario.}
    \label{Fig 3:constant delay}
\end{figure}

In ideal circumstances, RTD remains constant as it is solely determined by the transmission distance. 
Figure \ref{Fig 3:constant delay} shows a constant delay scenario, assuming an RTD-induced delay of $d_{RTD}=1$.
The state of the robot at the slave end evolves over time, while at the master end, due to the presence of RTD, the first observation is null, and subsequent observations lag behind those at the slave end by one time step, expressed as $x_t = s_{t-1} \delta(t>=1)+ \Phi \delta(t<1)$, where $\Phi$ indicates the null, and the $\delta$ function is the Dirac delta function.

\begin{figure}[t]
    \centering
    \includegraphics[width=0.6\textwidth]{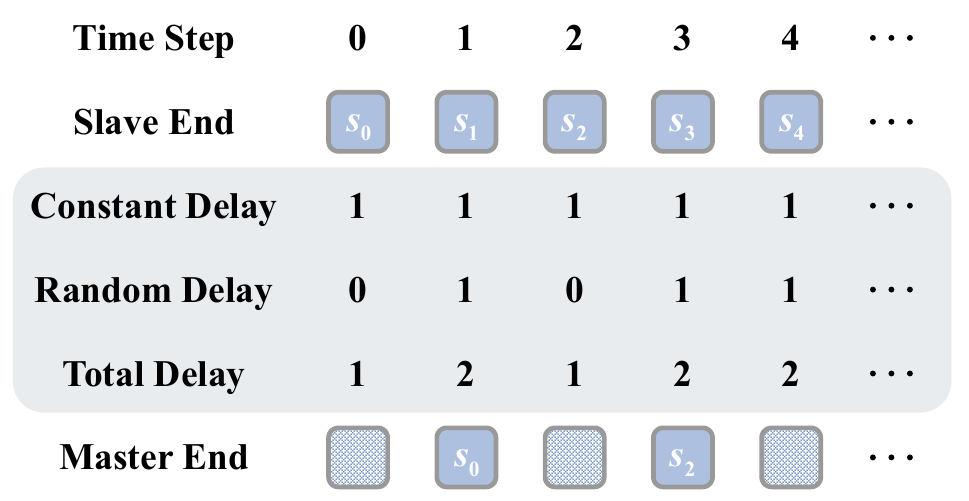}
    \caption{\ A random delay scenario.}
    \label{Fig 4:random delay}
\end{figure}

During the transmission, network congestion and other environmental factors often introduce randomness into the delay. 
Building upon the RTD, we assume that the random delay follows a uniform distribution with a parameter $\xi$ , denoted as $d_{random} \sim Uniform(\xi)$. 
Consequently, the total delay in data link is $d_{total}(t) = d_{RTD} + d_{random}(t)$.  
In Figure \ref{Fig 4:random delay},  a scenario with random delay is illustrated, where $d_{RTD}=1$ and $\xi=1$. 
At $t=0$, $d_{total}(0)=1$, indicating that the master end does not observe the state at $t=0$. Instead, it observes $s_0$ at $t=1$, resulting in $x_0=\Phi$ and $x_1=s_0$. 
Similarly, at $t=1$, $d_{total}(1)=2$, and at $t=2$, $d_{total}(2)=1$, implying that the master end does not observe the state at $t=2$, and both $s_1$ and $s_2$ are potentially observed at $t=3$. 
However, the old state is discarded, resulting in the observed state at $t=3$ being $s_2$, namely, $x_3=s_2$.
Note that, regardless of a constant or random delay, the focus of this study is on making decisions at each time step based on the observations from the master end.  The following sections will delve into a detailed analysis of this aspect.

\section{Method}
\label{sec:method}

The framework of space teleoperation based on DRL is illustrated in Figure \ref{Fig 2:spaceOverall Architecture}, consisting of three components: master end, data link, and slave end. 
The data link and slave end components operate in a manner akin to traditional teleoperation processes. 
The data link facilitates data transmission and processing, while the slave end primarily executes commands for the space robotic arm to interact with the environment and subsequently returns the state information and rewards generated by the environment.
The design of the master end diverges from traditional methods, incorporating a Delay Information Processing(DIP) module and a DRL module. 
Within the DIP module, we introduce three distinct methods: Mapping, Prediction, and State Augmentation. 
The DRL module predominantly utilizes the SAC algorithm.
The specific algorithm is outlined as follows.

\textbf{Remark}. The delay value at the master end is defined as follows:
\begin{equation}
    z_{t}= \begin{cases} z_{t-1} + 1,  & \text{if $x_t= \Phi$} ,\\ 
    d_{RTD}, & \text{others.} \end{cases}
\end{equation}

\subsection{The DIP module}
\label{Time delay Module}
\subsubsection{Mapping}
\begin{figure}[t]
    \centering
    \includegraphics[width=0.6\textwidth]{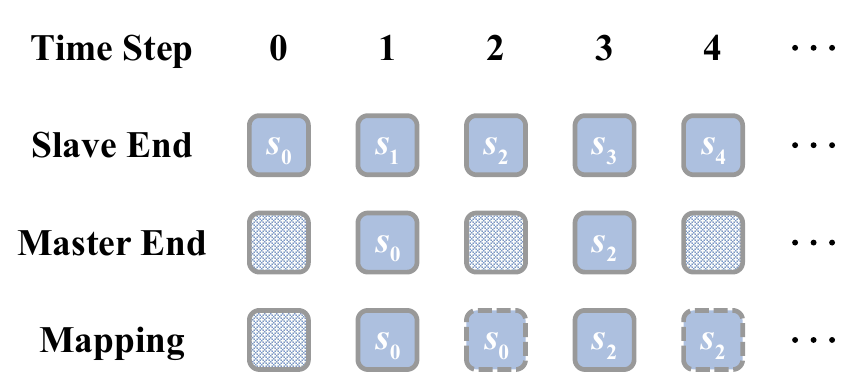 }
    \caption{\ A simple illustration of Mapping.}
    \label{Fig 5:mapping}
\end{figure}

Mapping adopts a memoryless strategy that disregards delays and treats the most recently observed state as the environment's true state for decision-making.
When $x_t$ differs from $\Phi$, the master end observes the current state; 
however, if $x_t$ equals $\Phi$, indicating an unobserved state at the master end, it is substituted with the last observed state $x_{t-z_t}$, and the instant reward $\hat{r}_t$ is set to 0.
Figure \ref{Fig 5:mapping} provides a simple illustrative example based on the random delay specified in Figure \ref{Fig 4:random delay}. 

For $t=1$ and $t=3$, the master end can directly observe $s_0$ and $s_2$, respectively, resulting in $x_1=s_0$ and $x_3=s_2$, with the instant reward calculated according to Eq. \ref{eq7}. 
By contrast, for  $t=2$ and $t=4$, the normal observation is unavailable, i.e., $x_2=x_4=\Phi$, and the current state is replaced with the last observed state, i.e., $\hat{x}_2=s_0$ and $\hat{x}_4=s_2$, while the corresponding instant reward is set to $0$. The algorithm is outlined in Algorithm \ref{alg:mapping}.

\begin{algorithm}[!h]
    \caption{Mapping}
    \label{alg:mapping}
    \renewcommand{\algorithmicrequire}{\textbf{Input:}}
    \renewcommand{\algorithmicensure}{\textbf{Output:}}
    \begin{algorithmic}[1]
        \REQUIRE action buffer $A$, current state at the master end $x_{t}$, current delay value $z_t$.  
        \ENSURE $x_{t}$ and reward $r_{t-1}$.    

        \IF {$x_t = \Phi$}
            \STATE $x_t = x_{t-z_t}$
            \STATE $r_{t-1} = 0$
        \ELSE
            \STATE $r_{t-1} = r(x_{t-1}, a_{t-1})$ using Eq. \ref{eq7}, where $a_{t-1}$ is from $A$
        \ENDIF

        \RETURN $x_{t}$ and $r_{t-1}$
    \end{algorithmic}
\end{algorithm}

\begin{figure}[h]
    \centering
    \includegraphics[width=0.8\textwidth]{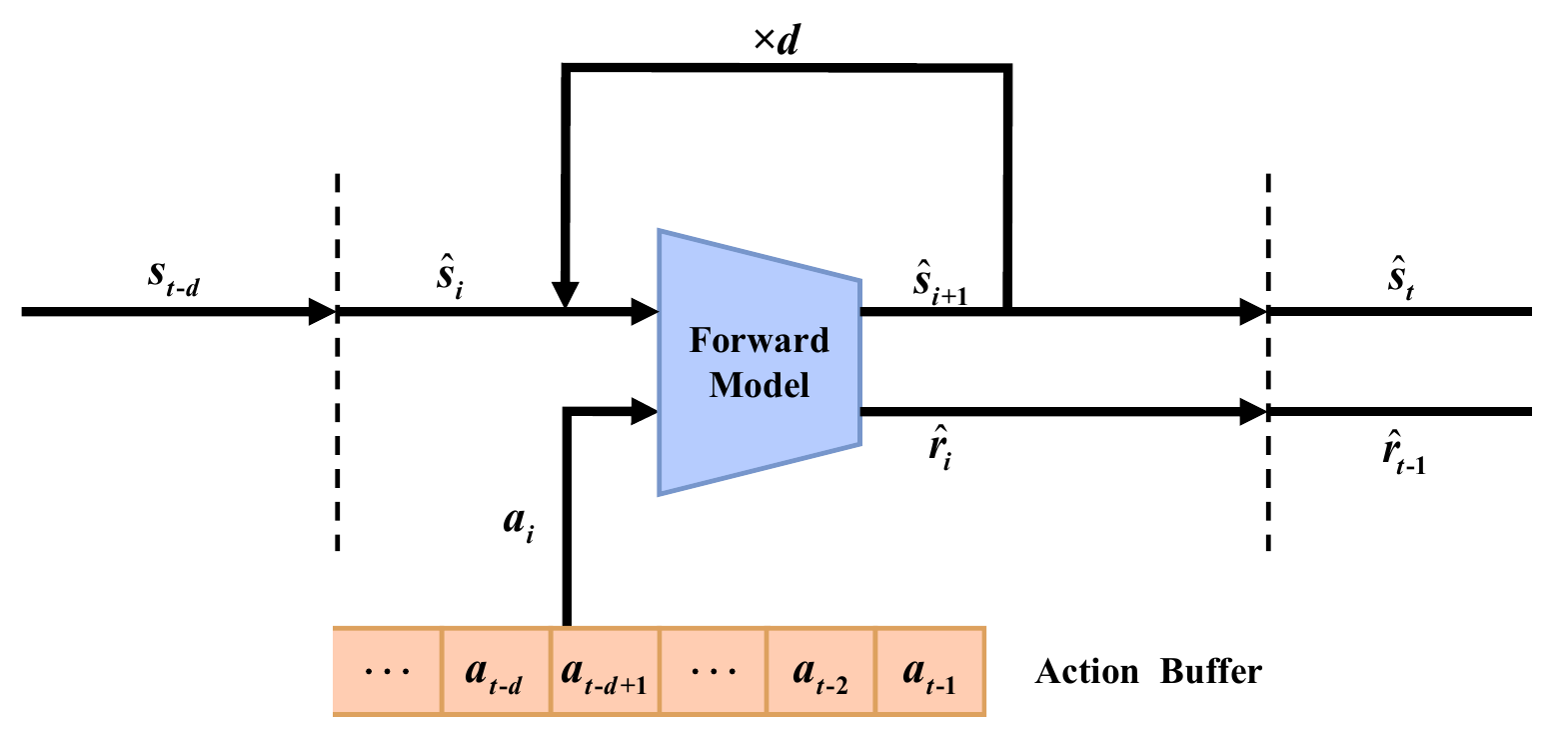 }
    \caption{\ A simple illustration of Prediction.}
    \label{Fig 6:prediction}
\end{figure}

\subsubsection{Prediction}
Prediction involves training a forward model using historical trajectory data.
To achieve this, we leverage the tuples $(s_t, a_t, r_t, s_{t+1}) $ stored in the Replay Buffer to train a nonlinear function $f$ through supervised learning. 
Here, the inputs are $s_t$ and $a_t$, while the outputs consist of $r_t$ and $s_{t+1}$, namely, $s_{t+1}, r_t = f(s_t, a_t)$, effectively modeling the dynamics of the environment.
Once the forward model is trained, we compute $\hat{x}_t$ and $\hat{r}_{t-1}$ iteratively using the last observed state $x_{t-z_t}$ and the historical action sequence $(a)_{k=t-z_t}^{k=t-1}$, as shown in Figure \ref{Fig 6:prediction}. 
The final decision is then made based on the predicted $\hat{x}_t$. The algorithm operates as Algorithm \ref{alg:prediction}.

\begin{algorithm}[!h]
    \caption{Prediction}
    \label{alg:prediction}
    \renewcommand{\algorithmicrequire}{\textbf{Input:}}
    \renewcommand{\algorithmicensure}{\textbf{Output:}}
    \begin{algorithmic}[1]
        \REQUIRE replay buffer $R$, action buffer $A$, current delay value $z_t$, current state at the master end $x_{t}$, forward model $f$, thresholds of steps for initiating the training of the forward model and retraining: $M$ and $N$.  
        \ENSURE $x_{t}$ and reward $r_{t-1}$.    

        \IF{len($R$) $ < M$}
            \STATE Execute Algorithm \ref{alg:mapping}
        \ELSE
            \STATE $s_{t-z_t} = x_t$
            \IF{len($R$) $ = M$ or len($R$) $ \% N = 0 $ }
               \STATE Train $f$ using dataset from $R$
            \ENDIF
            \FOR{each $k \in [t-{z_t},t-1]$}
                \STATE $s_{k+1}, r_{k} = f(s_k, a_k) $, where $a_k$ is from $A$
            \ENDFOR
            \STATE $x_t = s_t$
        \ENDIF
        \RETURN $x_{t}$ and $r_{t-1}$.
    \end{algorithmic}
\end{algorithm}

\subsubsection{State augmentation}
State augmentation constructs an Information State consisting of delayed state information and historical action sequences to transform the delayed MDP into a delay-free MDP, which is defined as follows:
\begin{itemize}[leftmargin=*]
    \item State $\mathscr{X}$. 
    
    For $t \ge d_{RTD}$, the new state  is defined as follows,
    \begin{equation}
    X_{t}= \begin{cases} (x_{t-z_t}, a_{t-d_{RTD}}, \cdots, a_{t-1}),  & \text{if $x_t= \Phi$} ,\\ 
    (x_t, a_{t-d_{RTD}}, \cdots, a_{t-1} ), & \text{others.} \end{cases}
\end{equation}
    \item Action $\mathscr{A}$. 
    
    It is the same as the definition in Section \ref{RL}.
    \item Reward $\mathscr{R}$. 
    
    Based on Eq. \ref{eq7}, the new reward function is as below:
    \begin{equation}
    r_{t}= \begin{cases} 0,  & \text{if $x_t= \Phi$} ,\\ 
    r(x_t, a_t), & \text{others.} \end{cases}
\end{equation}
    \item Initial state distribution $\rho$. 
    
    The new initial state distribution is as below:    
    \begin{equation}
    \rho(\mathscr{X}_0) = \rho(s_0)\prod \limits_{i=0}^{d_{RTD} - 1}{\delta(a_i - c_i)},
    \end{equation}
    where $\rho(s_0)$ is the original initial state distribution and $\{c_i\}_{i=0}^{d_{RTD} - 1}$ refers to the random action of the operator without the state from the remote.
\end{itemize}

The algorithm of delay information processing is as follows:

\begin{algorithm}[!h]
    \caption{State Augmentation}
    \label{alg:augmentation}
    \renewcommand{\algorithmicrequire}{\textbf{Input:}}
    \renewcommand{\algorithmicensure}{\textbf{Output:}}
    \begin{algorithmic}[1]
        \REQUIRE action buffer $A$, current delay value $z_t$, current state at the master end $x_{t}$.  
        \ENSURE $x_{t}$ and reward $r_{t-1}$.    

         \IF {$x_t = \Phi$}
            \STATE $x_t = (x_{t-z_t}, a_{t-d_{RTD}}, \cdots, a_{t-1})$, where $\{a_i\}_{t-d_{RTD}}^{t-1}$ is from $A$
            \STATE $r_{t-1} = 0$
        \ELSE
            \STATE $x_t = (x_t, a_{t-d_{RTD}}, \cdots, a_{t-1} )$
            \STATE $r_{t-1} = r(x_{t-1}, a_{t-1})$ using Eq. \ref{eq7}, where $\{a_i\}_{t-d_{RTD}}^{t-1}$ is from $A$
        \ENDIF
        \RETURN $x_{t}$ and $r_{t-1}$
    \end{algorithmic}
\end{algorithm}

\subsection{The decision module}
The objective function  at the master end is designed as follows:
\begin{equation}
    \mathcal{J}(\pi) = \sum_{t=0}^{T}{
    \mathbb{E}_{(x_t, a_t)\sim D}[r(x_t, a_t)]
    }, \label{eq12}
\end{equation}
where $x_t$ is the delayed state after processing in Section \ref{Time delay Module} and $D$ is the distribution of previously sampled states and actions, or a replay buffer.
The optimal policy is obtained by maximizing the objective function. 
However, the end-to-end trajectory tracking for space robots poses a significant challenge due to its complex nature and the expansive state space dimension.
For example, for Algorithm \ref{alg:mapping} and Algorithm \ref{alg:prediction}, the dimension is $2n+19$ or $2n+31$, while the dimension extends to $2n+19 +  n  d_{RTD}$ or $2n+31+  n  d_{RTD}$ for Algorithm \ref{alg:augmentation}.
Furthermore, optimizing Eq. \ref{eq12}  yields a deterministic policy and often leads to local optima, making it difficult to find policy parameters that achieve high cumulative rewards.
To address this, Eq. \ref{eq12} is modified as follows:
\begin{equation}
    \mathcal{J}(\pi) = \sum_{t=0}^{T}{
    \mathbb{E}_{(x_t, a_t)\sim D}[r(x_t, a_t) + \alpha\mathcal{H}(\pi(\cdot|x_t))]
    }, \label{eq13}
\end{equation}
where $\mathcal{H}(\pi(\cdot|x_t))$ represents the entropy of the policy, and $\alpha$ determines the relative importance of the entropy term against the reward. 
Eq. \ref{eq13} not only encourages the agent to explore more effectively but also controls the stochasticity of the optimal policy.

In this study, Eq. \ref{eq13} is optimized  based on the SAC algorithm. 
Similar to traditional SAC algorithms, three neural networks $Q_{\theta}{(x_t, a_t)}$, $V_{\psi}(x_t)$ and $\pi_{\phi}(a_t|x_t)$ are designed to to approximate the state-action value function, state value function, and policy function, respectively.
The parameters $\theta$ of the Q-function are updated by minimizing the soft Bellman residual:
\begin{equation}
    \mathcal{J}_{Q}(\theta)=\mathbb{E}_{(x_t, a_t)\sim D}{[\frac{1}{2}(Q_{\theta}(x_t, a_t) - r(x_t, a_t) - \gamma\mathbb{E}_{x_{t+1}\sim\mathscr{P}}{[V_{\overline{\psi}}{(x_{t+1})}]})^2]} \label{eq14}
\end{equation}
where $V_{\overline{\psi}}$ represents the target value network for 
$V_{\psi}$  and is updated by the exponential moving average of the value network weights. 
The parameters $\psi$ of the soft value function are updated by minimizing the squared residual error:
\begin{equation}
    \mathcal{J}_{V}(\psi) = \mathbb{E}_{x_t \sim D}{[\frac{1}{2}(
    V_{\psi}(x_t) - \mathbb{E}_{a_t\sim \pi_{\phi}}{[Q_{\theta}(x_t, a_t) - log\pi_{\phi}(a_t|x_t)]^2}
    )]}. \label{eq15}
\end{equation}
The parameters $\phi$ of the policy network are updated by minimizing the KL-divergence:
\begin{equation}
    \mathcal{J}_{\pi}(\phi)=\mathbb{E}_{x_t \sim D}[
    D_{KL}(\pi_{\phi}(\cdot|x_t) || \frac{exp(Q_{\theta}(x_t, \cdot))}{Z_{\theta}(x_t)})
    ],\label{eq16}
\end{equation}
where $Z_{\theta}(x_t)$ is the partition function and can be ignored.
We utilize the reparameterization trick to minimize Eq. \ref{eq16} and construct the policy network:
\begin{equation}
    a_t = f_{\phi}(\epsilon_t;x_t),
\end{equation}
where $\epsilon_t$ is a noise vector. Consequently, Eq. \ref{eq16} can be rewritten as:
\begin{equation}
    \mathcal{J}_{\pi}(\phi)=\mathbb{E}_{x_t \sim D, \epsilon_t \sim \mathcal{N}}[
    log\pi_{\phi}(f_\phi(\epsilon_t;x_t)|x_t) - Q_\theta(x_t, f_\phi(\epsilon_t;x_t))
    ],\label{eq18}
\end{equation}
The only difference from the traditional SAC lies in structure of the replay buffer. 
In traditional methods, a quadruple is stored in the replay buffer after each step.
However, in teleoperation, the master end may not have access to the true state. 
To ensure the reliability of the data stored in the replay buffer, such data cannot be included.

The pseudocode of the complete procedure is specified in Algorithm \ref{alg:teleoperation}.

\begin{algorithm}[!h]
    \caption{Teleoperation based on SAC}
    \label{alg:teleoperation}
    \renewcommand{\algorithmicrequire}{\textbf{Input:}}
    \renewcommand{\algorithmicensure}{\textbf{Output:}}
    \begin{algorithmic}[1]
        \REQUIRE replay buffer $R$, action buffer $A$, current delay value $z_t$.  
        \ENSURE policy $\pi$.    

        \STATE $z_t=d_{RTD}$
        \FOR{each iteration}
            \FOR{each $t \in [0, d_{RDT} - 1]$}    
                \STATE $a_t \sim random(-a_{max}, a_{max})$
                \STATE $A.append(a_t)$
            \ENDFOR
            \STATE
            \FOR{each $ t \in [d_{RDT}, T]$}
                \STATE obtain $x_t$ and $r_{t-1}$ from the slave end 
                
                \IF{$x_t = \Phi$}
                    \STATE $flag = true, z_t \gets z_t + 1$
                \ELSE
                    \STATE $flag = false, z_t \gets d_{RTD}$
                \ENDIF
                \STATE get new $x_t$ using Algorithm \ref{alg:mapping}, Algorithm \ref{alg:prediction} or Algorithm \ref{alg:augmentation} 
                \STATE $a_t \sim \pi_{\Phi}(\cdot|x_t)$
                \STATE obtain $x_{t+1}$ and $r_t$ from the slave end 
                \STATE  get new $x_{t+1}$ using Algorithm \ref{alg:mapping}, Algorithm \ref{alg:prediction} or Algorithm \ref{alg:augmentation} 
                \IF{$x_t \neq \Phi$}
                    \STATE $R \gets  R \cup \{ (x_t, a_{t-z_t}, r_{t-1}, x_t)\}$
                \ENDIF
                
            \ENDFOR
            \STATE update all parameters by applying gradient descent on Eq. \ref{eq14}, Eq. \ref{eq15} and Eq. \ref{eq18}.
        \ENDFOR
        
    \end{algorithmic}
\end{algorithm}

\section{Experiments}
\label{sec:experiment}
\subsection{Simulation settings}

We introduce the simulation settings from three aspects: environment setup, neural network architecture, and operating platform.

\begin{figure}[h]
    \centering
    \includegraphics[width=0.5\textwidth]{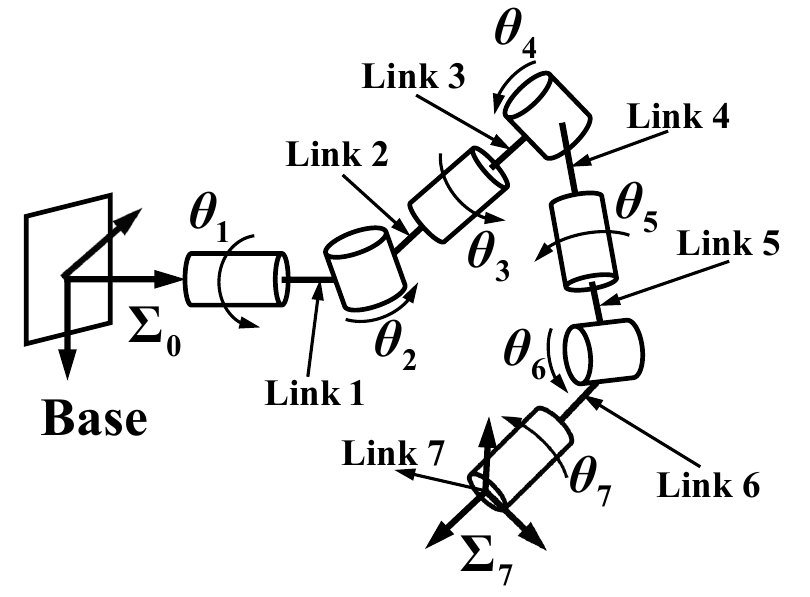}
    \caption{\ 7-DOF redundant manipulator.}
    \label{fig 7:7-DOF redundant manipulator}
\end{figure}

\begin{figure}[h]
    \centering
    \includegraphics[width=0.5\textwidth]{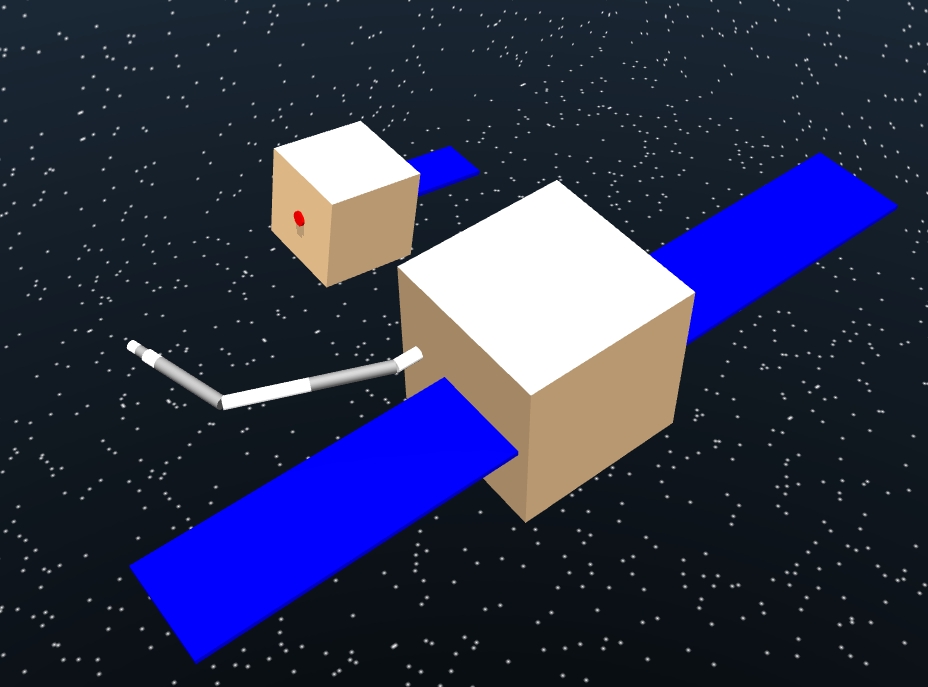}
    \caption{\ Space robot in MuJoCo.}
    \label{fig 8:Space robot in MuJoCo}
\end{figure}

\textbf{Environment setup.}    
Based on the 7-DOF redundant robotic arm model\cite{wu2020reinforcement}, illustrated in Figure \ref{fig 7:7-DOF redundant manipulator}, we construct a single-arm space robot model within the MuJoCo environment, as shown in Figure \ref{fig 8:Space robot in MuJoCo}. 
The kinematic and dynamic parameters are detailed in Table \ref{tab: kinematic}.
In the constructed simulation environment, the length of each simulation time step $T_m$ is set to 0.01 s, with an action executed every 4 time steps.
Throughout the experiments, the maximum number of environment steps per episode is $T=250$, resulting in a maximum simulation duration of $T \cdot T_m \cdot 4 = 10$ seconds. This experiment involves a random grasping task performed in teleoperation mode, aiming to position the end effector of the space manipulator within 5 mm of the target point. Given the end effector's radius of 0.02 m, we set $d_{threshold}=0.025$  m, as outlined in Table \ref{tab: kinematic}.

\begin{table}[H]
\footnotesize
\setlength{\abovecaptionskip}{0.1cm}  
\caption{Kinematic and dynamic properties.}
\centering
\begin{tabular}{lll}
\toprule
\textbf{Link No.}&\textbf{Shape}&\textbf{Inertial properties} \\
\midrule
\textbf{0(Base)}&Box,\ Edge length 0.7m&$m=200$\ kg \\
&&$\boldsymbol{I}=diag(40,35,100)\ \mathrm{kg\cdot m^2}$ \\
\textbf{1}&Cylinder,\ $r=0.025\ \mathrm{m},$&$m=0.39$\ kg \\
&$l=0.1\ \mathrm{m}$&$\boldsymbol{I}=diag(3.9,3.9,1.2)\times 10^{-4}\ \mathrm{kg\cdot m^2}$ \\
\textbf{2}&Cylinder,\ $r=0.025\ \mathrm{m},$&$m=1.18$\ kg \\
&$l=0.3\ \mathrm{m}$&$\boldsymbol{I}=diag(160,160,3.7)\times 10^{-4}\ \mathrm{kg\cdot m^2}$ \\
\textbf{3}&Cylinder,\ $r=0.025\ \mathrm{m},$&$m=1.18$\ kg \\
&$l=0.3\ \mathrm{m}$&$\boldsymbol{I}=diag(160,160,3.7)\times 10^{-4}\ \mathrm{kg\cdot m^2}$ \\
\textbf{4}&Cylinder,\ $r=0.025\ \mathrm{m},$&$m=1.18$\ kg \\
&$l=0.3\ \mathrm{m}$&$\boldsymbol{I}=diag(160,160,3.7)\times 10^{-4}\ \mathrm{kg\cdot m^2}$ \\
\textbf{5}&Cylinder,\ $r=0.025\ \mathrm{m},$&$m=0.24$\ kg \\
&$l=0.06\ \mathrm{m}$&$\boldsymbol{I}=diag(11,7.4,7.4)\times 10^{-5}\ \mathrm{kg\cdot m^2}$ \\
\textbf{6}&Cylinder,\ $r=0.02\ \mathrm{m},$&$m=0.1$\ kg \\
&$l=0.04\ \mathrm{m}$&$\boldsymbol{I}=diag(2.3,2.3,2)\times 10^{-5}\ \mathrm{kg\cdot m^2}$ \\
\textbf{7}&Cylinder,\ $r=0.02\ \mathrm{m},$&$m=0.1$\ kg \\
&$l=0.04\ \mathrm{m}$&$\boldsymbol{I}=diag(2.3,2.3,2)\times 10^{-5}\ \mathrm{kg\cdot m^2}$ \\
\bottomrule
\label{tab: kinematic}
\end{tabular}
\end{table}

\begin{table}[H]
\footnotesize
\setlength{\abovecaptionskip}{0.1cm}  
\caption{Different simulation environments.}
\centering
\begin{tabular}{lll}
\toprule
\textbf{Env}&\textbf{Base}&\textbf{Target} \\
\midrule
\textbf{SAFBFT}&Fixed&Fixed \\
\textbf{SAFBRT}&Fixed&Rotated,  $torque \sim U(0,1)$ or $torque\sim U(-1,0)$ \\
\textbf{SAUBFT}&Free-floating&Fixed \\
\textbf{SAUBRT}&Free-floating&Rotated,$torque \sim U(0,1)$ or $torque\sim U(-1,0)$  \\
\bottomrule
\label{tab: environments}
\end{tabular}
\end{table}

Based on the model in Figure \ref{fig 8:Space robot in MuJoCo}, we built four distinct simulation environments, each corresponding to a specific scenario, provided in Table \ref{tab: environments}.

To enhance the model's generalization, we conduct uniform sampling within a cube in the workspace range of the robotic arm and treat them as the positions of target during training. 
Additionally, we introduce noise to the initial joint angles and angular velocities of the robotic arm. 
Specifically, in the world coordinate system, the position of the base is 
\begin{equation}
    p_b = [0, 0, 1],
\end{equation}
and the position of the target star is 
\begin{equation}
    p_s = [-1.05 \pm 0.1, -0.125 \pm 0.1, 1 \pm 0.1].
\end{equation}
The initial joint angles of the robotic arm without noise are
\begin{equation}
    q_0 = [0, \frac{\pi}{7}, 0, \frac{2\pi}{3}, 0, 0, 0],
\end{equation}
and the noise added to the initial angles is represented as 
\begin{equation}
    \begin{aligned}
        q_{noise} = [&U(-\frac{\pi}{8}, \frac{\pi}{8}), U(-\frac{\pi}{8}, \frac{\pi}{8}), U(-\frac{\pi}{8}, \frac{\pi}{8}), U(-\frac{\pi}{8}, \frac{\pi}{8}), \\ &U(-\frac{\pi}{6}, \frac{\pi}{6}), U(-\frac{\pi}{6}, \frac{\pi}{6}), U(-\frac{\pi}{6}, \frac{\pi}{6})].
    \end{aligned}
\end{equation}
Similarly, the initial angular velocities of the robotic arm without noise 
are
\begin{equation}
    v_0 = [0, 0, 0, 0, 0, 0, 0],
\end{equation}
and the noise added to the initial velocities is denoted as 
\begin{equation}
    \begin{aligned}
        v_{noise} = [&U(-0.005, 0.005), U(-0.005, 0.005), U(-0.005, 0.005), \\ &U(-0.005, 0.005), U(-0.005, 0.005), U(-0.005, 0.005), \\ & U(-0.005, 0.005)].
    \end{aligned}
\end{equation}
Consequently, the initial joint angles of the robotic arm are given by 
\begin{equation}
q = q_0 + q_{noise},
\end{equation}
and the initial angular velocities are given by 
\begin{equation}
    v = v_0 + v_{noise}.
\end{equation}
The termination condition for each episode is either reaching the maximum number of simulation steps $T$ or the distance between the end effector and the tracking target point being less than or equal to $d_{threshold}$.

The hyperparameters $\omega_1, \cdots, \omega_5$ in Eq. \ref{eq7}  are set as follows: -0.01, -1, -0.1, -0.1, and 1.5, respectively.

\textbf{Neural network architecture.}
The network architectures and hyperparameters of the generic SAC algorithm used in this paper are presented in Table \ref{tab: SAC}.

\begin{table}[h]
\footnotesize
\caption{The network architectures and hyperparameters of the generic SAC algorithm.}
\begin{tabular}{c|ccll}
\hline
\multirow{16}{*}{\makecell{\bf{Structure of} \\ \bf{SAC}}}
& \multicolumn{4}{c}{\bf{Actor Network}}                                           
   \\ \cline{2-5} 
   & \multirow{4}{*}{dim of input} 
   & \multicolumn{1}{c}{ \multirow{3}{*}{dim of  state} } & Alg. \ref{alg:mapping} or \ref{alg:prediction} & {33 or 45}                                   
   \\
   &  &   & Alg. \ref{alg:augmentation} & \makecell[l]{33+ 7 * $d_{RTD}$  or \\ 45 + 7 * $d_{RTD}$} 
   \\   \cline{3-5}                                            
     &  &  \multicolumn{2}{c}{dim of action}  & 7                           
     \\  \cline{2-5} 
   & \multicolumn{3}{c}{dim of hidden size}   & 256                           
     \\
   & \multicolumn{3}{c}{activation function} & Relu                             \\
   & \multicolumn{3}{c}{number of layers}   & 2                                  \\
    & \multicolumn{3}{c}{dim of output}    & 7                                  \\ \cline{2-5}
    & \multicolumn{4}{c}{\bf{Critc Network}}
       \\ \cline{2-5} 
   & \multirow{4}{*}{dim of input} 
   & \multicolumn{1}{c}{ \multirow{3}{*}{dim of  state} } & Alg. \ref{alg:mapping} or \ref{alg:prediction} & {33 or 45}                                   
   \\
   &  &   & Alg. \ref{alg:augmentation} & \makecell[l]{33+ 7 * $d_{RTD}$  or \\ 45 + 7 * $d_{RTD}$} 
   \\   \cline{3-5}                                            
     &  &  \multicolumn{2}{c}{dim of action}  & 7                           
     \\  \cline{2-5} 
    & \multicolumn{3}{c}{dim of hidden size}   & 256                          \\
    & \multicolumn{3}{c}{activation function}   & Relu                        \\
    & \multicolumn{3}{c}{number of layers}  & 2                                \\
    & \multicolumn{3}{c}{dim of output}  & 1                                    \\ \cline{1-5}
    \multirow{8}{*}{\makecell{\bf{Hyper}-\\ \bf{parameters}}}  
    & \multicolumn{3}{c}{\makecell[c]{$\gamma$}}  & 0.99                       \\
    & \multicolumn{3}{c}{learning rate}    & 0.0003                            \\
    & \multicolumn{3}{c}{\makecell[c]{$\alpha$}}    & 0.2                      \\
    & \multicolumn{3}{c}{batch size} & 256                                     \\
    & \multicolumn{3}{c}{total training step}    & 500000                      \\
    & \multicolumn{3}{c}{replay buffer size}        & 500000                   \\
    & \multicolumn{3}{c}{optimizer}      & Adam                               \\
    & \multicolumn{3}{c}{\makecell[c]{$\tau$}}    & 0.005                      \\ \cline{1-5}
\end{tabular}
\label{tab: SAC}
\end{table}

\begin{table}[h]
\footnotesize
\caption{the network architecture and parameters of the forward model.}
\centering
\begin{tabular}{c|ccl}
\hline
\multirow{7}{*}{\makecell{\bf{Forward Moel}}}
& \multicolumn{3}{c}{\bf{Network}}                           
   \\ \cline{2-4} 
   & \multirow{2}{*}{dim of input} 
   & \multicolumn{1}{c}{ \multirow{1}{*}{dim of  state} } &  {33 or 45}                              
   \\                                           
     &  &  \multicolumn{1}{c}{dim of action}  & 7                           
     \\  \cline{2-4} 
   & \multicolumn{2}{c}{dim of hidden size}   & 200                           
     \\
   & \multicolumn{2}{c}{activation function} & Relu                             \\
   & \multicolumn{2}{c}{number of layers}   & 3                                  \\
    & \multicolumn{2}{c}{dim of output}    & {33 or 45}                                  
    \\ \cline{1-4}                               
    \multirow{3}{*}{\makecell{\bf{Hyper}-\\ \bf{parameters}}}  
    & \multicolumn{2}{c}{\makecell[c]{capacity of training}}  & 50000                      \\
    & \multicolumn{2}{c}{optimizer}    & Adam         \\
    & \multicolumn{2}{c}{learning rate} & 0.001  
    \\ \cline{1-4}
\end{tabular}
\label{tab: forward}
\end{table}

When using Algorithm \ref{alg:prediction}, the network architecture and parameters of the forward model are presented in Table \ref{tab: forward}.

\textbf{Operating platform.}
All experiments are conducted on an NVIDIA GeForce RTX 3090 graphics card. The versions of gym, mujoco, and PyTorch used in the experiments are 0.21.0, 2.0.2.8, and 1.11, respectively.

\subsection{Simulation result and analysis}

\begin{figure}[h]
    \centering
    \includegraphics[width=0.95\textwidth]{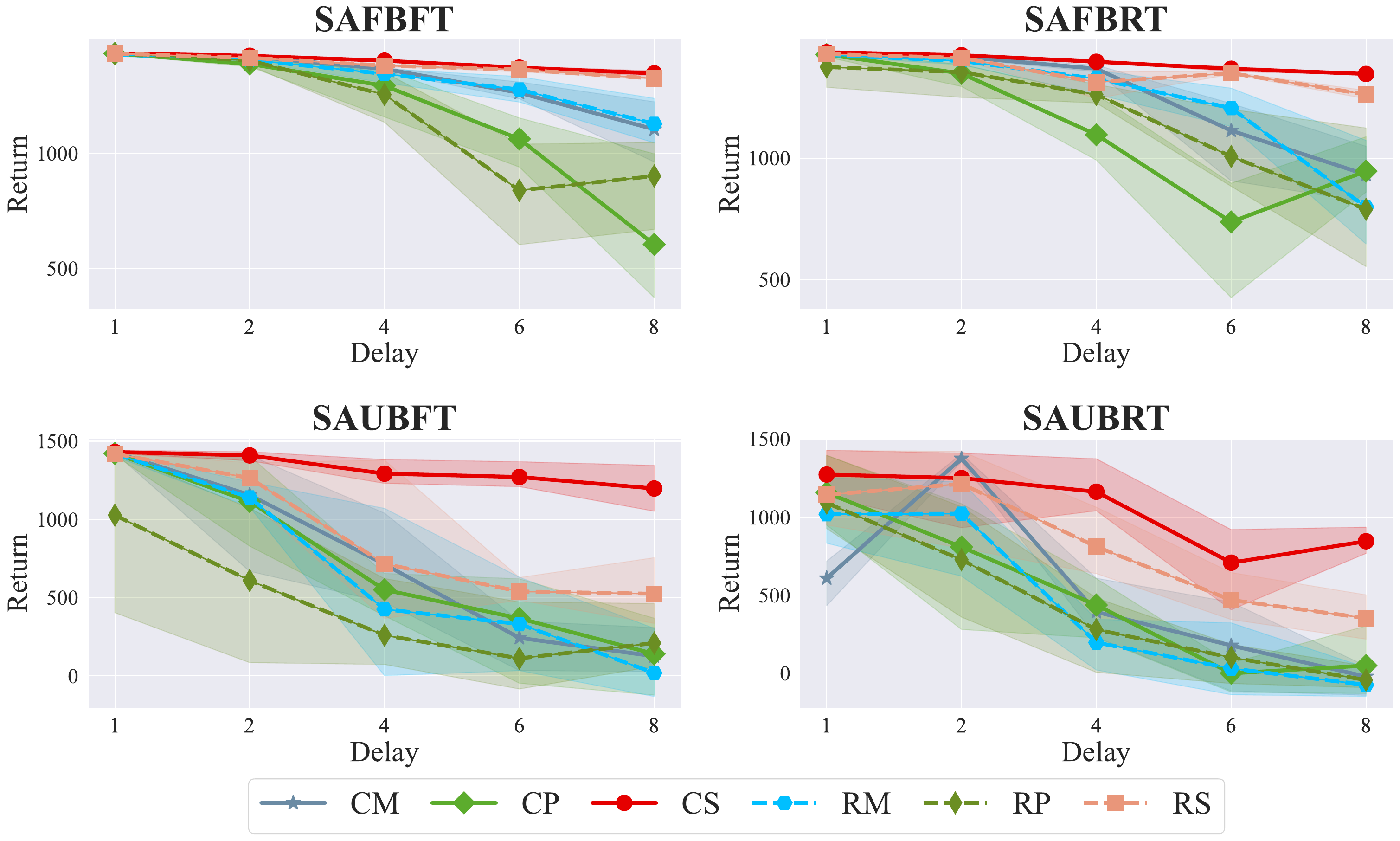 }
    \caption{\ Comparison of different algorithms across various scenarios.}
    \label{Fig 9:final}
\end{figure}


For the four environments in Table \ref{tab: environments}, we conduct experiments with fixed delays $d_{RTD}=1,2,4,6,8$ and corresponding random delays with $\xi=2$ for each algorithm. 
The state augmentation method based on SAC is named as SACAS. 
We abbreviate constant\_Mapping as CM, and similarly constant\_Prediction, constant\_SACAS, random\_Mapping, random\_Prediction, and random\_SACAS as CP, CS, RM, RP, and RS, respectively. 
Each data point in Figure \ref{Fig 9:final}, \ref{Appendix:cm} - \ref{Appendix:rs}, and every entry in Tables \ref{tab: delay=1}-\ref{tab: delay=8} are obtained from experiments with three different random seeds.

\subsubsection{Investigation of the training process}
Observations from Figure \ref{Fig 9:final} include:
1) In SAFBFT, the curves of CS and RS exhibit consistent patterns, as do those of CM and RM, indicating that whether the delay is random has minimal impact on the algorithms, with CS and RS consistently outperforming CM and RM. 
The curves of CP and RP decrease significantly with increasing delays.
2) In SAFBRT, CS and RS exhibit the highest rankings in performance, although the effectiveness of SACAS experiences fluctuations due to random delays, as evident in the RS curve.
Following closely are the CM and RM curves, with RM slightly edging out CM particularly at a delay of 6. 
Bringing up the rear are CP and RP.
3) In SAUBFT, CS demonstrates superior performance than others, with the RS curve maintaining a higher standing than the rest, irrespective of a slight but discernible variation with increasing delay. 
CM, RM, and CP initially perform comparably at delay 1, but their effectiveness notably declines with increasing delay. In contrast, RP noticeably lags behind other curves in terms of performance.
4) In SAUBRT, CS and RS continue to maintain their lead, although with a declining trend. 
The other curves show significant decline, except for CM, which performs far better than in other environments at delay 2, possibly attributed to its decision state.

Based on the above observations, we can draw the following conclusions:
\begin{itemize}[leftmargin=*]
\item In the four environments, the performance of all three algorithms deteriorates as the delay increases, regardless of whether it is fixed or random. Fixed delay yields better results than random delay with the same $d_{RTD}$.
\item SACAS consistently demonstrates the highest and most stable performance, except in the most challenging scenario SAUBRT, where its random performance is less remarkable compared to other environments, but is still superior overall.
\item Mapping is the second-best performer, showing commendable performance in both fixed-base scenarios and capable of handling situations where the base is floating and the delay is small.
\item Prediction performs worst across all environments, exhibiting the most severe performance decline as delay increases. This can be possibly attributed to the substantial discrepancies between the predicted states produced by the forward model over successive iterations and the actual states upon which the decisions are based.
\end{itemize}

Figure \ref{Appendix:cm} - \ref{Appendix:rs} illustrate the training records of 50,000 environment steps under various delay conditions for the three algorithms. 
It is evident that the curves of SACAS exhibit greater density, with turning points leaning further to the upper left, indicating faster convergence and smoother target tracking. 
Furthermore, SACAS performs better in scenarios with free-floating space, rotating targets, and random delays compared with the other two methods.

\subsubsection{Investigation of the evaluation process}
After training,  we evaluate the trained models for 100 test runs within their respective training scenarios, and record the success rates achieved and the total steps taken upon success, as shown in Tables \ref{tab: delay=1}-\ref{tab: delay=8}.

From Tables \ref{tab: delay=1} and \ref{tab: delay=2}, it is evident that when confronted with low delays ($d_{RTD}=1, 2$) and operating within a fixed-base space robot environment, irrespective of target rotation or random delays, all three algorithms exhibit comparable success rates. 
However, in scenarios such as SAUBFT with a floating base, Predication demonstrates notably lower performance compared with the others. Similarly, SAUBRT, Mapping exhibits less favorable outcomes than SACAS and Prediction, while Predication requires more steps to achieve success.
Table \ref{tab: delay=4} reveals that when moderate delays ($d_{RTD}=4$) are introduced, in the SAFBFT scenario, the success rates of the three algorithms are comparable, but Prediction requires more steps than the others. 
In the SAFBRT scenario, both Mapping and SACAS demonstrate similar task completion efficiency, while Prediction initially lags behind. 
In SAUBFT and SAUBRT, with fixed delays, both Mapping and Prediction achieve less than half of SACAS's success rate, while SACAS maintains a success rate of over 90\%. 
Under random delays, the situation exacerbates, but SACAS still achieves a success rate close to 70\%.
Tables \ref{tab: delay=6} and \ref{tab: delay=8} further elucidate that when confronted with large delays ($d_{RTD}=6, 8$) and operating within a fixed-base space robot framework, both Prediction and Mapping achieve a minimum success rate of no less than 60\%. 
However, under floating base conditions, both algorithms may exhibit poor performance, potentially even reaching a success rate of 0\%. 
For SACAS, under fixed base conditions, the success rate surpasses 96\%, while the success rate remains consistently above 50\% when the target is fixed under floating base conditions and it does not drop below 33\% when the target rotates.

In summary, SACAS method demonstrates broad applicability across various scenarios, while Mapping is effective particularly in fixed-base scenarios. However, owing to its requirement to forecast delay states, Predication tends to accumulate accumulates errors in iterative predictions, leading to notable deviations from the true values, thereby impacting decision-making processes, and hence is primarily suitable for scenarios characterized by low delays.

\begin{table}[h]
\footnotesize
\caption{Comparison of success rates and step counts across various algorithms in different environments over one hundred episodes within $d_{RTD}=1$}
\scalebox{0.94}{
\begin{tabular}{clllll}
\toprule
\multirow{2}{*}{\textbf{Env}} & \multicolumn{1}{c}{\multirow{2}{*}{\textbf{Algorithm}}} & \multicolumn{2}{c}{\textbf{constant}} & \multicolumn{2}{c}{\textbf{random}}   
\\ \cline{3-6}
& \multicolumn{1}{c}{}   & \textbf{success rate} & \textbf{episode steps} & \textbf{success rate} & \textbf{episode steps} 
\\  \midrule
\multirow{3}{*}{\textbf{SAFBFT}}  
& Mapping & 1.0 ± 0.0    & 13.65 ± 2.91  & 1.0 ± 0.0    & 15.13 ± 3.66  
\\ & Prediction   & 1.0 ± 0.0    & 14.5 ± 4.04   & 1.0 ± 0.0    & 15.71 ± 5.78  
\\ & SACAS    & 1.0 ± 0.0    & 13.58 ± 2.83  & 1.0 ± 0.0    & 14.45 ± 2.97 
\\ \midrule
\multirow{3}{*}{\textbf{SAFBRT}}  
& Mapping   & 1.0 ± 0.0    & 14.53 ± 4.02  & 1.0 ± 0.0    & 16.25 ± 4.35  
\\ & Prediction   & 0.99 ± 0.0   & 15.12 ± 4.43  & 0.99 ± 0.0   & 17.16 ± 5.95  
\\ & SACAS   & 1.0 ± 0.0    & 13.44 ± 2.79  & 1.0 ± 0.0    & 15.57 ± 4.73  
\\ \midrule
\multirow{3}{*}{\textbf{SAUBFT}}  
& Mapping    & 1.0 ± 0.0    & 14.75 ± 3.55  & 0.96 ± 0.03  & 21.02 ± 7.86 
\\ & Prediction & 0.96 ± 0.05  & 17.51 ± 6.05  & 0.88 ± 0.11  & 19.88 ± 11.01 
\\ & SACAS   & 0.99 ± 0.01  & 14.86 ± 4.61  & 0.97 ± 0.01  & 17.6 ± 4.72  
\\ \midrule
\multirow{3}{*}{\textbf{SAUBRT}}  
& Mapping & 0.55 ± 0.12  & 23.89 ± 17.27 & 0.89 ± 0.04  & 23.68 ± 9.85  
\\ & Prediction  & 0.96 ± 0.31  & 34.36 ± 29.69 & 0.91 ± 0.04  & 23.03 ± 10.76 
\\ & SACAS & 0.94 ± 0.08  & 16.11 ± 5.42  & 0.94 ± 0.06  & 19.24 ± 7.38 
\\ \bottomrule
\label{tab: delay=1}
\end{tabular}
}
\end{table}

\begin{table}[!h]
\footnotesize
\caption{Comparison of success rates and step counts across various algorithms in different environments over one hundred episodes within $d_{RTD}=2$}
\scalebox{0.94}{
\begin{tabular}{clllll}
\toprule
\multirow{2}{*}{\textbf{Env}} & \multicolumn{1}{c}{\multirow{2}{*}{\textbf{Algorithm}}} & \multicolumn{2}{c}{\textbf{constant}} & \multicolumn{2}{c}{\textbf{random}}   
\\ \cline{3-6}
& \multicolumn{1}{c}{}   & \textbf{success rate} & \textbf{episode steps} & \textbf{success rate} & \textbf{episode steps} 
\\  \midrule
\multirow{3}{*}{\textbf{SAFBFT}} 
& Mapping & 1.0 ± 0.0    & 16.88 ± 4.24  & 0.99 ± 0.0   & 18.86 ± 6.0  
\\ & Prediction   & 0.99 ± 0.01  & 22.87 ± 13.57 & 1.0 ± 0.0    & 18.86 ± 6.74  
\\ & SACAS    & 0.99 ± 0.01  & 15.33 ± 2.97  & 1.0 ± 0.0    & 16.94 ± 4.43 
\\ \midrule
\multirow{3}{*}{\textbf{SAFBRT}} 
& Mapping   & 1.0 ± 0.0    & 17.7 ± 5.36   & 1.0 ± 0.0    & 22.44 ± 9.67  
\\ & Prediction    & 0.99 ± 0.01  & 25.37 ± 19.99 & 0.99 ± 0.01  & 24.38 ± 13.79
\\ & SACAS   & 1.0 ± 0.0    & 16.19 ± 4.41  & 1.0 ± 0.0    & 17.88 ± 4.21  
\\ \midrule
\multirow{3}{*}{\textbf{SAUBFT}} 
& Mapping    & 0.97 ± 0.03  & 21.27 ± 8.2   & 0.85 ± 0.05  & 25.54 ± 9.94
\\ & Prediction & 0.77 ± 0.14  & 28.01 ± 13.76 & 0.73 ± 0.09  & 25.45 ± 8.53
\\ & SACAS   & 0.94 ± 0.04  & 18.9 ± 6.66   & 0.96 ± 0.04  & 19.52 ± 5.72   
\\ \midrule
\multirow{3}{*}{\textbf{SAUBRT}} 
& Mapping & 0.93 ± 0.04  & 22.82 ± 7.56  & 0.8 ± 0.11   & 26.88 ± 10.09 
\\ & Prediction  & 0.95 ± 0.18  & 30.54 ± 22.99 & 0.82 ± 0.05  & 32.64 ± 14.84 
\\ & SACAS & 0.92 ± 0.11  & 22.03 ± 14.09 & 0.9 ± 0.05   & 21.46 ± 11.27
\\ \bottomrule
\label{tab: delay=2}
\end{tabular}
}
\end{table}

\begin{table}[!h]
\footnotesize
\caption{Comparison of success rates and step counts across various algorithms in different environments over one hundred episodes within $d_{RTD}=4$}
\scalebox{0.94}{
\begin{tabular}{clllll}
\toprule
\multirow{2}{*}{\textbf{Env}} & \multicolumn{1}{c}{\multirow{2}{*}{\textbf{Algorithm}}} & \multicolumn{2}{c}{\textbf{constant}} & \multicolumn{2}{c}{\textbf{random}}   
\\ \cline{3-6}
& \multicolumn{1}{c}{}   & \textbf{success rate} & \textbf{episode steps} & \textbf{success rate} & \textbf{episode steps} 
\\  \midrule
\multirow{3}{*}{\textbf{SAFBFT}} 
& Mapping & 1.0 ± 0.0    & 24.68 ± 8.78  & 1.0 ± 0.0    & 30.92 ± 16.09 
\\ & Prediction   & 1.0 ± 0.0    & 39.24 ± 28.78 & 0.98 ± 0.03  & 43.57 ± 29.38  
\\ & SACAS     & 1.0 ± 0.0    & 19.63 ± 2.96  & 1.0 ± 0.0    & 23.28 ± 8.31 
\\ \midrule
\multirow{3}{*}{\textbf{SAFBRT}} 
& Mapping   & 0.99 ± 0.01  & 28.9 ± 14.51  & 1.0 ± 0.0    & 35.06 ± 22.51 
\\ & Prediction    & 0.83 ± 0.1   & 57.1 ± 46.38  & 0.95 ± 0.02  & 43.3 ± 31.74
\\ & SACAS   & 1.0 ± 0.0    & 20.44 ± 4.68  & 0.98 ± 0.01  & 29.22 ± 23.53  
\\ \midrule
\multirow{3}{*}{\textbf{SAUBFT}} 
& Mapping    & 0.45 ± 0.19  & 27.57 ± 11.2  & 0.45 ± 0.22  & 33.4 ± 17.18
\\ & Prediction & 0.42 ± 0.09  & 39.18 ± 22.88 & 0.28 ± 0.08  & 32.76 ± 12.41
\\ & SACAS   & 0.98 ± 0.01  & 22.55 ± 5.71  & 0.65 ± 0.2   & 26.39 ± 6.24  
\\ \midrule
\multirow{3}{*}{\textbf{SAUBRT}} 
& Mapping & 0.41 ± 0.11  & 34.2 ± 23.2   & 0.26 ± 0.07  & 39.06 ± 24.25
\\ & Prediction  & 0.48 ± 0.05  & 48.68 ± 26.42 & 0.33 ± 0.12  & 40.48 ± 22.39 
\\ & SACAS & 0.92 ± 0.07  & 27.91 ± 9.24  & 0.7 ± 0.06   & 33.53 ± 15.37
\\ \bottomrule
\label{tab: delay=4}
\end{tabular}
}
\end{table}

\begin{table}[!h]
\footnotesize
\caption{Comparison of success rates and step counts across various algorithms in different environments over one hundred episodes within $d_{RTD}=6$}
\scalebox{0.94}{
\begin{tabular}{clllll}
\toprule
\multirow{2}{*}{\textbf{Env}} & \multicolumn{1}{c}{\multirow{2}{*}{\textbf{Algorithm}}} & \multicolumn{2}{c}{\textbf{constant}} & \multicolumn{2}{c}{\textbf{random}}   
\\ \cline{3-6}
& \multicolumn{1}{c}{}   & \textbf{success rate} & \textbf{episode steps} & \textbf{success rate} & \textbf{episode steps} 
\\  \midrule
\multirow{3}{*}{\textbf{SAFBFT}} 
& Mapping & 0.99 ± 0.0   & 44.97 ± 29.48 & 0.99 ± 0.01  & 48.31 ± 33.55
\\ & Prediction   & 0.87 ± 0.01  & 76.61 ± 53.1  & 0.87 ± 0.05  & 58.48 ± 44.69  
\\ & SACAS     & 1.0 ± 0.0    & 24.04 ± 3.28  & 1.0 ± 0.0    & 28.3 ± 10.27
\\ \midrule
\multirow{3}{*}{\textbf{SAFBRT}} 
& Mapping   & 0.97 ± 0.0   & 48.97 ± 29.99 & 0.95 ± 0.03  & 57.43 ± 39.0  
\\ & Prediction    & 0.68 ± 0.11  & 83.5 ± 62.9   & 0.82 ± 0.02  & 70.2 ± 49.65
\\ & SACAS   & 1.0 ± 0.0    & 24.35 ± 3.91  & 0.99 ± 0.01  & 34.24 ± 21.69  
\\ \midrule
\multirow{3}{*}{\textbf{SAUBFT}} 
& Mapping    & 0.42 ± 0.11  & 39.95 ± 19.44 & 0.39 ± 0.17  & 34.45 ± 10.58
\\ & Prediction & 0.43 ± 0.09  & 51.14 ± 32.13 & 0.35 ± 0.1   & 66.68 ± 47.01
\\ & SACAS   & 0.90 ± 0.42  & 29.16 ± 8.62  & 0.57 ± 0.11  & 31.89 ± 7.64  
\\ \midrule
\multirow{3}{*}{\textbf{SAUBRT}} 
& Mapping & 0.2 ± 0.15   & 36.3 ± 17.67  & 0.22 ± 0.12  & 39.15 ± 17.13
\\ & Prediction  & 0.21 ± 0.06  & 49.04 ± 19.84 & 0.13 ± 0.06  & 46.95 ± 27.22 
\\ & SACAS & 0.72 ± 0.05  & 37.52 ± 14.88 & 0.42 ± 0.1   & 36.71 ± 15.56
\\ \bottomrule
\label{tab: delay=6}
\end{tabular}
}
\end{table}

\begin{table}[!h]
\footnotesize
\caption{Comparison of success rates and step counts across various algorithms in different environments over one hundred episodes within $d_{RTD}=8$}
\scalebox{0.94}{
\begin{tabular}{clllll}
\toprule
\multirow{2}{*}{\textbf{Env}} & \multicolumn{1}{c}{\multirow{2}{*}{\textbf{Algorithm}}} & \multicolumn{2}{c}{\textbf{constant}} & \multicolumn{2}{c}{\textbf{random}}   
\\ \cline{3-6}
& \multicolumn{1}{c}{}   & \textbf{success rate} & \textbf{episode steps} & \textbf{success rate} & \textbf{episode steps} 
\\  \midrule
\multirow{3}{*}{\textbf{SAFBFT}} 
& Mapping & 0.98 ± 0.01  & 61.97 ± 41.78 & 0.93 ± 0.0   & 69.96 ± 47.04
\\ & Prediction   & 0.56 ± 0.13  & 99.4 ± 60.81  & 0.47 ± 0.27  & 94.89 ± 69.96  
\\ & SACAS    & 1.0 ± 0.0    & 28.5 ± 4.32   & 1.0 ± 0.0    & 33.36 ± 12.15
\\ \midrule
\multirow{3}{*}{\textbf{SAFBRT}} 
& Mapping   & 0.87 ± 0.02  & 70.59 ± 49.27 & 0.78 ± 0.02  & 76.88 ± 53.64  
\\ & Prediction    & 0.81 ± 0.09  & 81.2 ± 52.83  & 0.6 ± 0.2    & 79.91 ± 51.07
\\ & SACAS   & 0.99 ± 0.0   & 29.71 ± 8.42  & 0.96 ± 0.02  & 43.52 ± 25.03  
\\ \midrule
\multirow{3}{*}{\textbf{SAUBFT}} 
& Mapping   & 0.17 ± 0.09  & 43.87 ± 20.37 & 0.12 ± 0.1   & 56.49 ± 29.58
\\ & Prediction & 0.16 ± 0.07  & 56.81 ± 18.75 & 0.17 ± 0.0   & 48.66 ± 21.69
\\ & SACAS   & 0.95 ± 0.03  & 35.4 ± 8.53   & 0.5 ± 0.16   & 37.48 ± 9.13   
\\ \midrule
\multirow{3}{*}{\textbf{SAUBRT}} 
& Mapping & 0.13 ± 0.09  & 57.32 ± 40.68 & 0.04 ± 0.02  & 50.92 ± 20.69
\\ & Prediction  & 0.15 ± 0.06  & 70.9 ± 46.8   & 0.03 ± 0.02  & 56.4 ± 41.99
\\ & SACAS & 0.74 ± 0.11  & 38.29 ± 18.25 & 0.38 ± 0.05  & 36.17 ± 10.8
\\ \bottomrule
\label{tab: delay=8}
\end{tabular}
}
\end{table}

\subsubsection{Performance comparison}
\begin{figure}[t]
    \centering
    \includegraphics[width=0.7\textwidth]{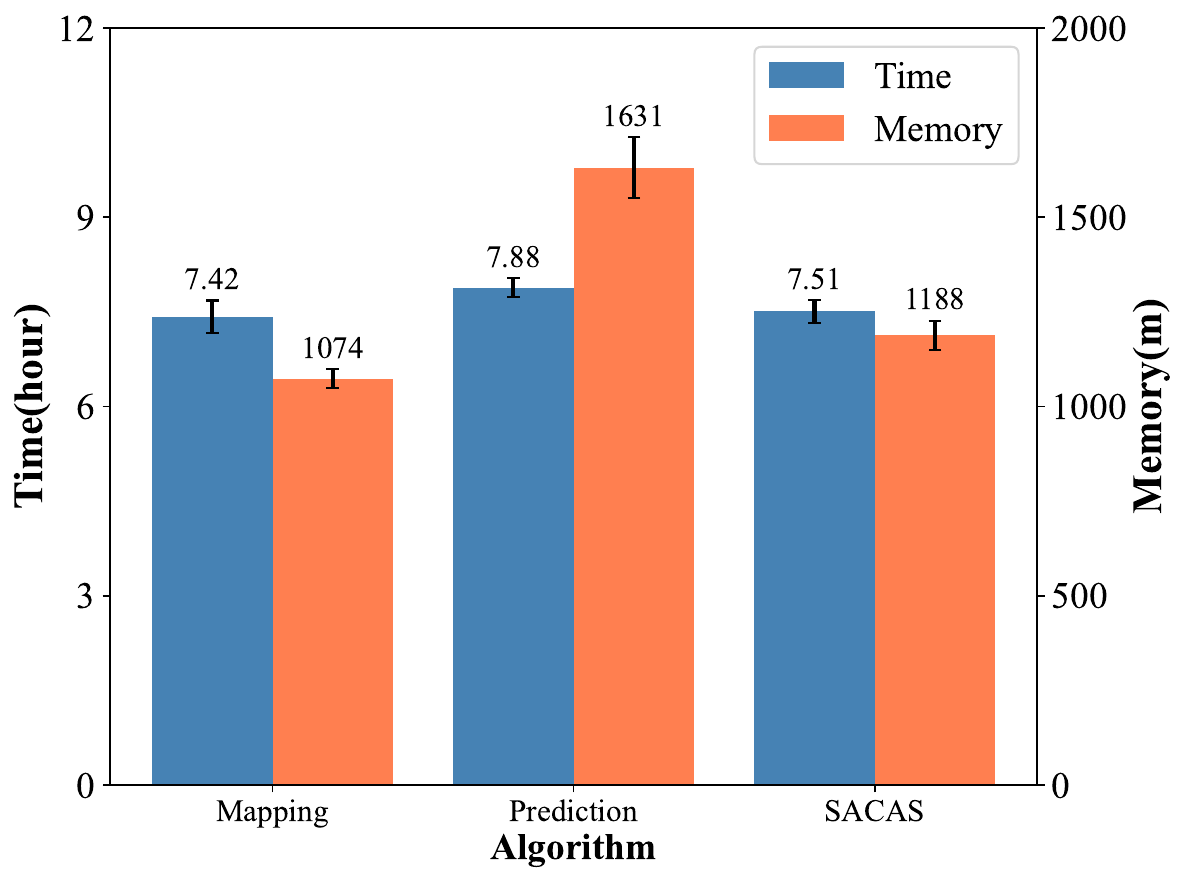 }
    \caption{\ Performance comparison of different algorithms.}
    \label{fig:preformance}
\end{figure}

We record the time and memory consumption of different algorithms across various environments during training and calculate the mean and variance of time and memory consumption for each algorithm, as shown in the Figure \ref{fig:preformance}.
Figure \ref{fig:preformance} illustrates Mapping stands out as the most efficient option in terms of both time and space utilization. 
SACAS, due to its expanded state space, necessitates larger network structures compared with Mapping, resulting in higher memory consumption and longer computation time. 
Although Prediction maintains a consistent state space, its periodic updates to the forward dynamics model significantly extend both time and memory consumption. 
Consequently, from the aforementioned analysis, it becomes evident that SACAS emerges as the optimal algorithm, effectively balancing effectiveness and performance.

\section{Conclusion}
\label{sec:conclusion}
In this study, we endeavor to tackle the intricate challenge of trajectory planning for teleoperated space manipulators. Our approach integrates deep reinforcement learning into the conventional telecontrol framework, representing the first instance of such innovative attempts in this field.
Our method focuses on the utilization of delayed state information and historical action buffers. To achieve this, we introduce three innovative techniques: Mapping, Prediction, and State Augmentation. These methods are designed to capitalize on delayed state information and historical actions to augment the decision-making capabilities of the agent, ensuring resilience even in environments characterized by inherent delays.
Through experimentation conducted across four distinct environments within the MuJoCo simulation environment, wherein we varied the fixity of both the base and target, we validated the effectiveness of these algorithms. Furthermore, our results highlight the superior efficiency and robustness of the State Augmentation method.

Future work could be pursued along three avenues: 1)The Prediction method, commonly employed in engineering, could be further enhanced through ensemble techniques to predict multi-step trajectories, thereby reducing cumulative errors, akin to the approach described in \cite{liu2021probabilistic}. 
2) The scope of investigation could be broadened to encompass more complex systems, such as multi-arm space manipulator models or flexible space arms, building upon the foundation of the single-arm rigid body model. 
3)The research could tackle more challenging problems, such as spatial activities involving force interactions, to further push the boundaries of inquiry and advance our understanding of teleoperated space manipulators.

\appendix

\section{More details on training process}
\label{sec:sample:appendix}
\begin{figure}[!h]
    \centering
    \includegraphics[width=0.95\textwidth]{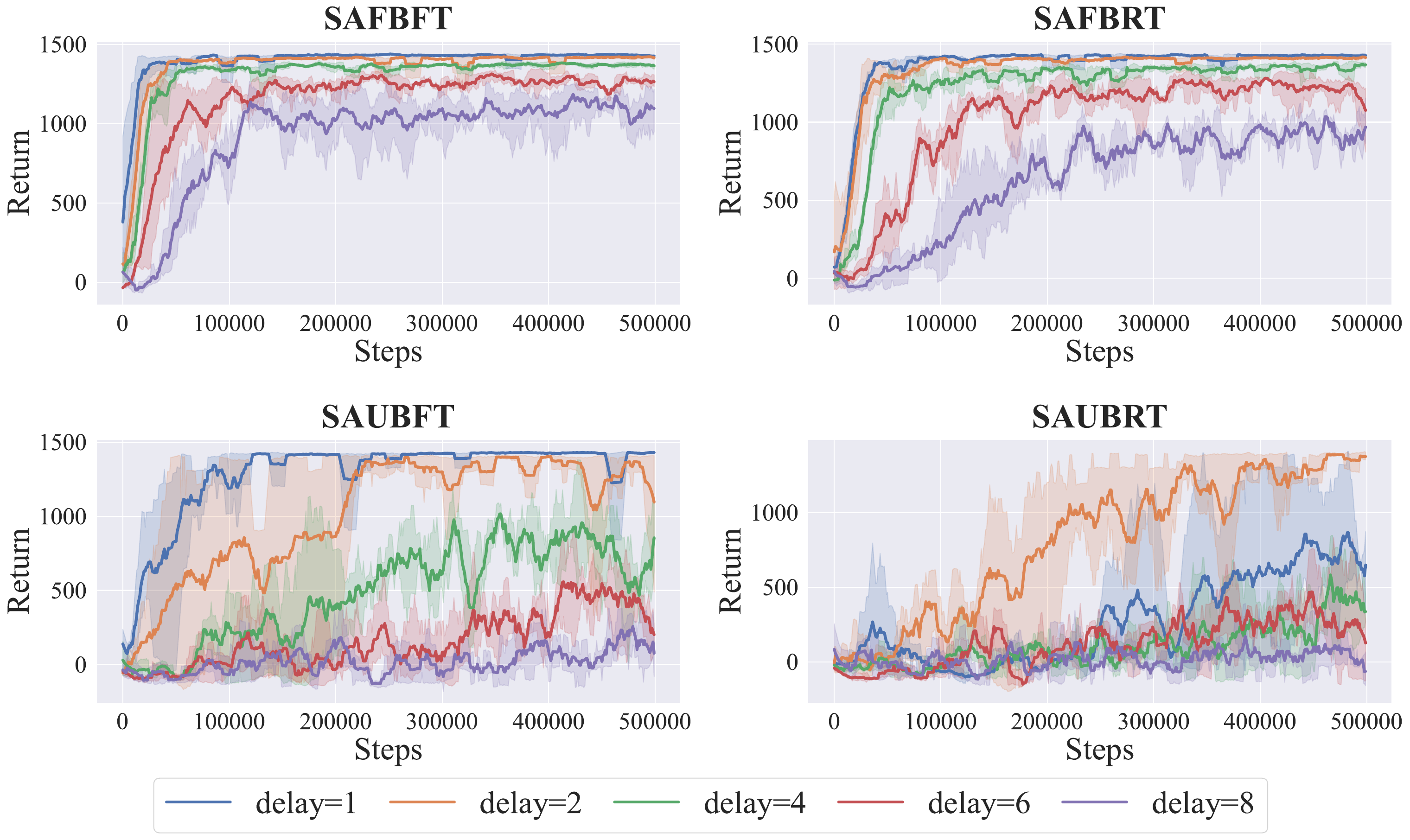 }
    \caption{\ Comparison of training with Mapping for different tasks under constant delay.}
    \label{Appendix:cm}
\end{figure}

\begin{figure}[!h]
    \centering
    \includegraphics[width=0.95\textwidth]{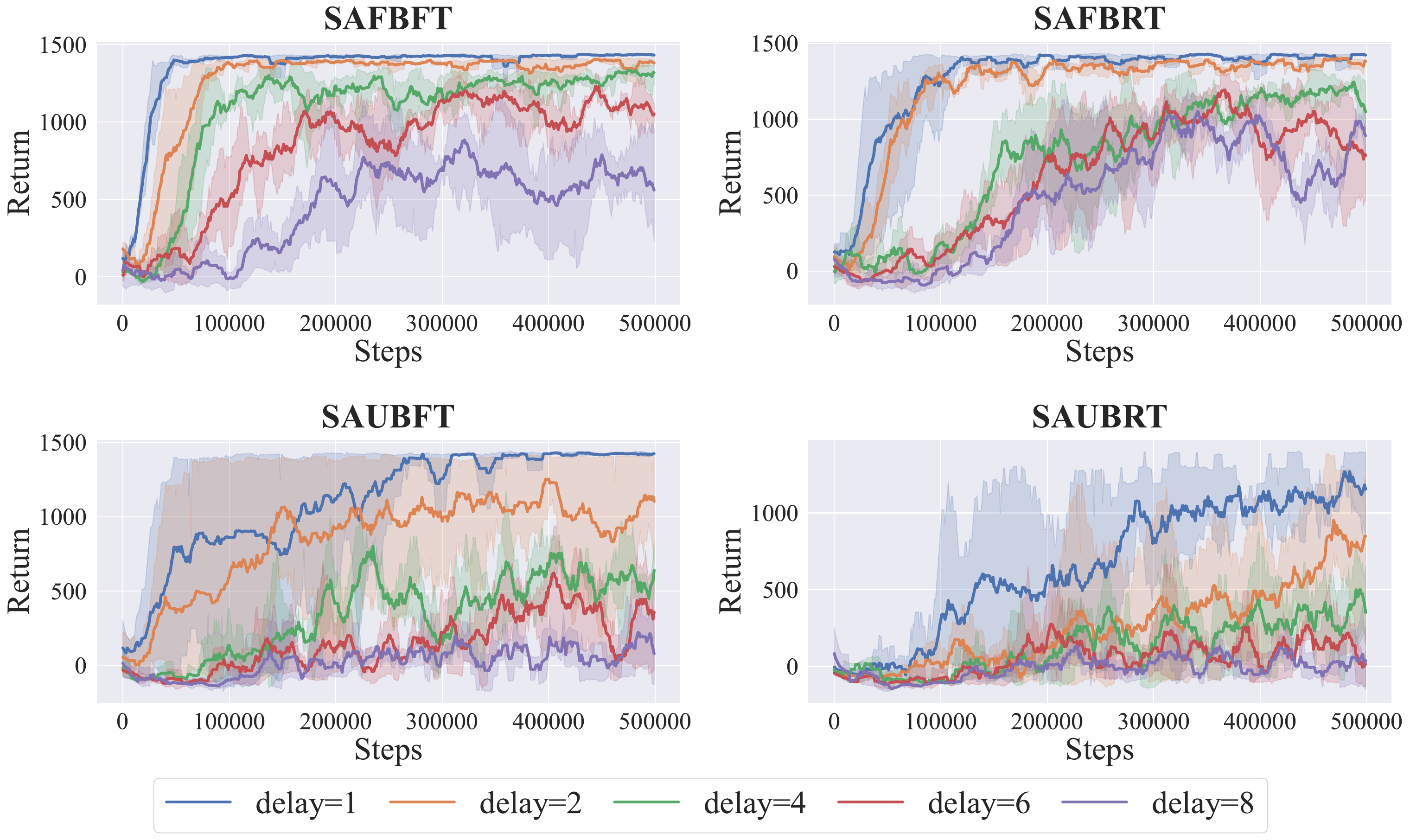 }
    \caption{\ Comparison of training with Prediction for different tasks under constant delay.}
    \label{Appendix:cp}
\end{figure}

\begin{figure}[!h]
    \centering
    \includegraphics[width=0.95\textwidth]{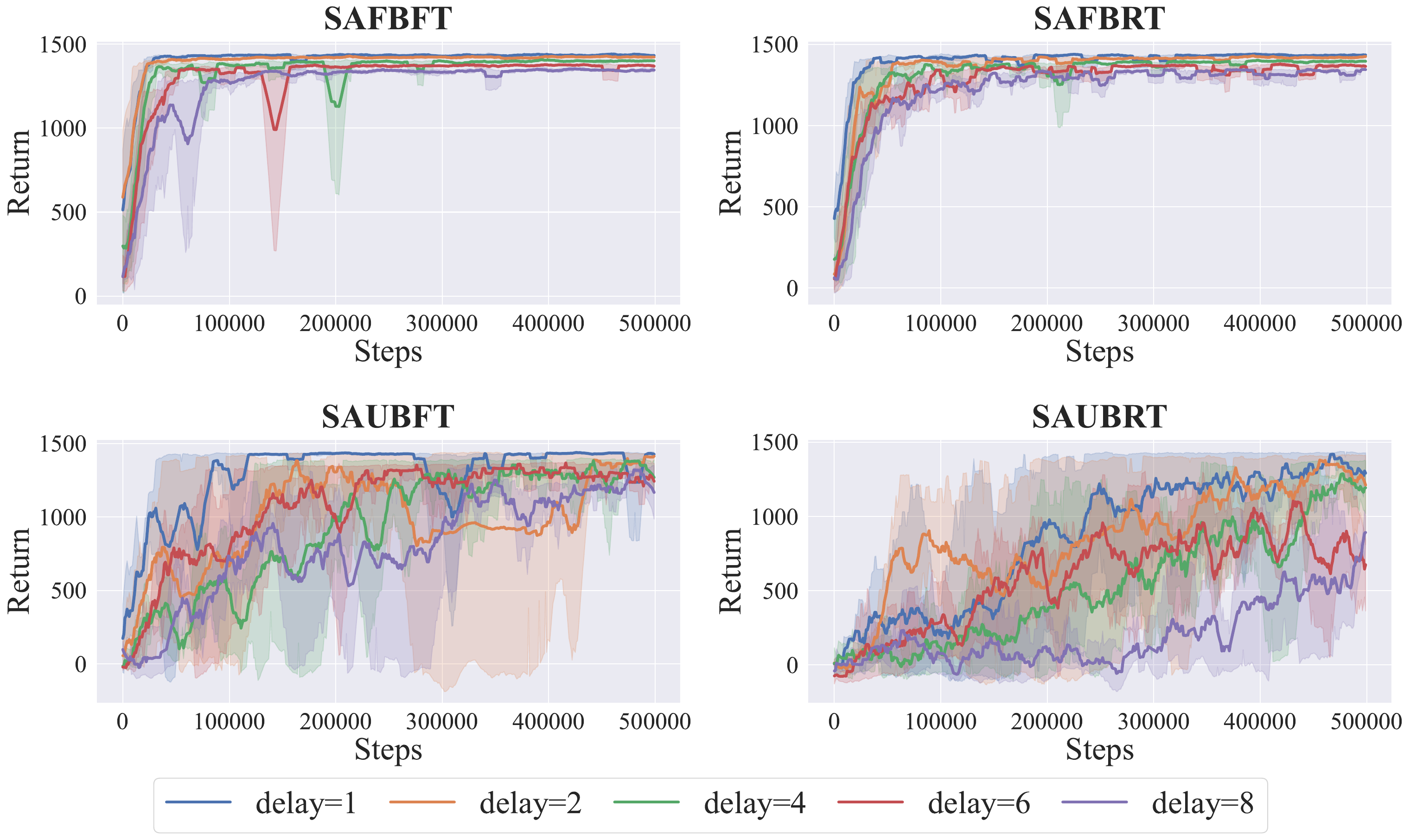 }
    \caption{\ Comparison of training with SACAS for different tasks under constant delay.}
    \label{Appendix:cs}
\end{figure}

\begin{figure}[!h]
    \centering
    \includegraphics[width=0.95\textwidth]{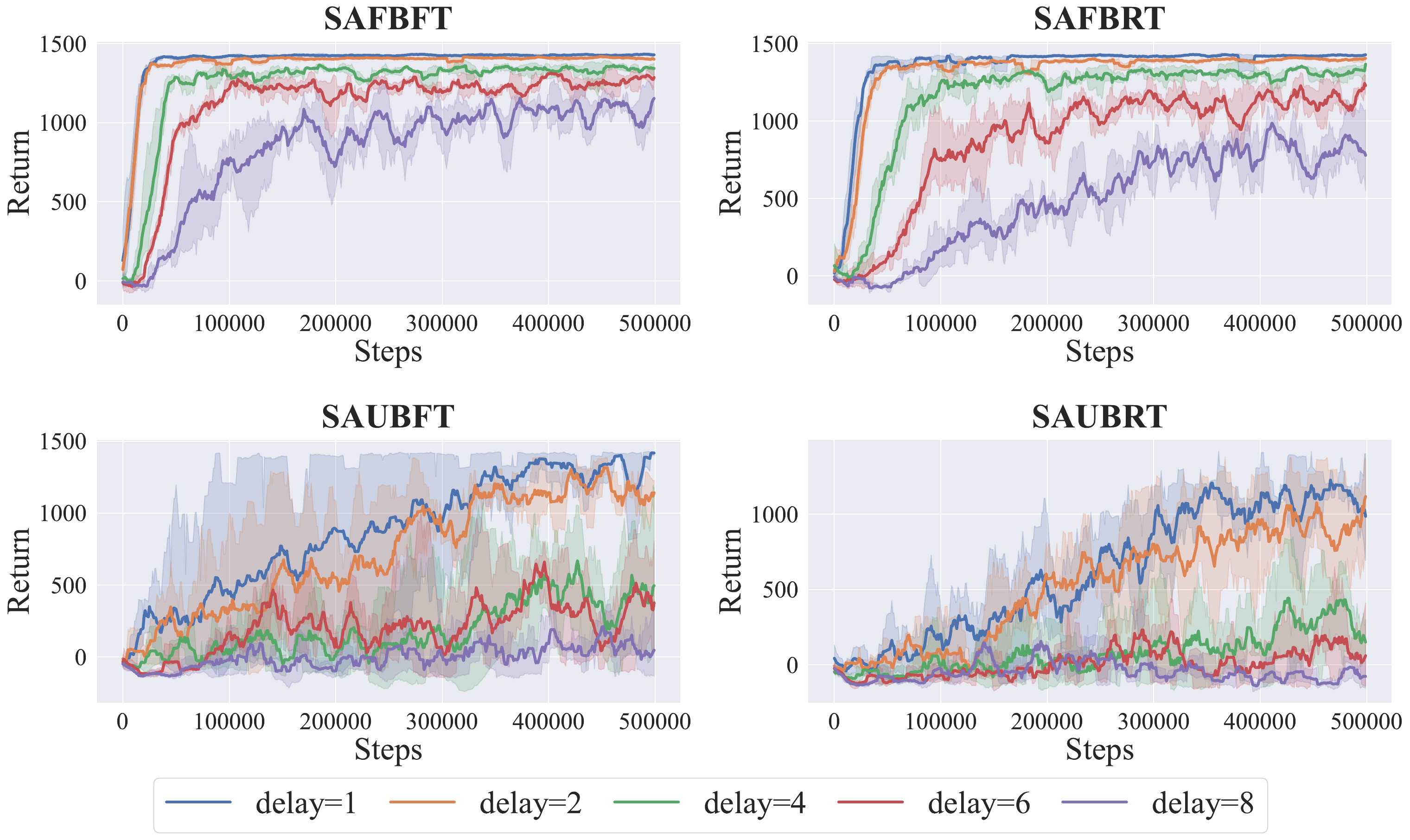 }
    \caption{\ Comparison of training with Mapping for different tasks under random delay.}
    \label{Appendix:rm}
\end{figure}

\begin{figure}[!h]
    \centering
    \includegraphics[width=0.95\textwidth]{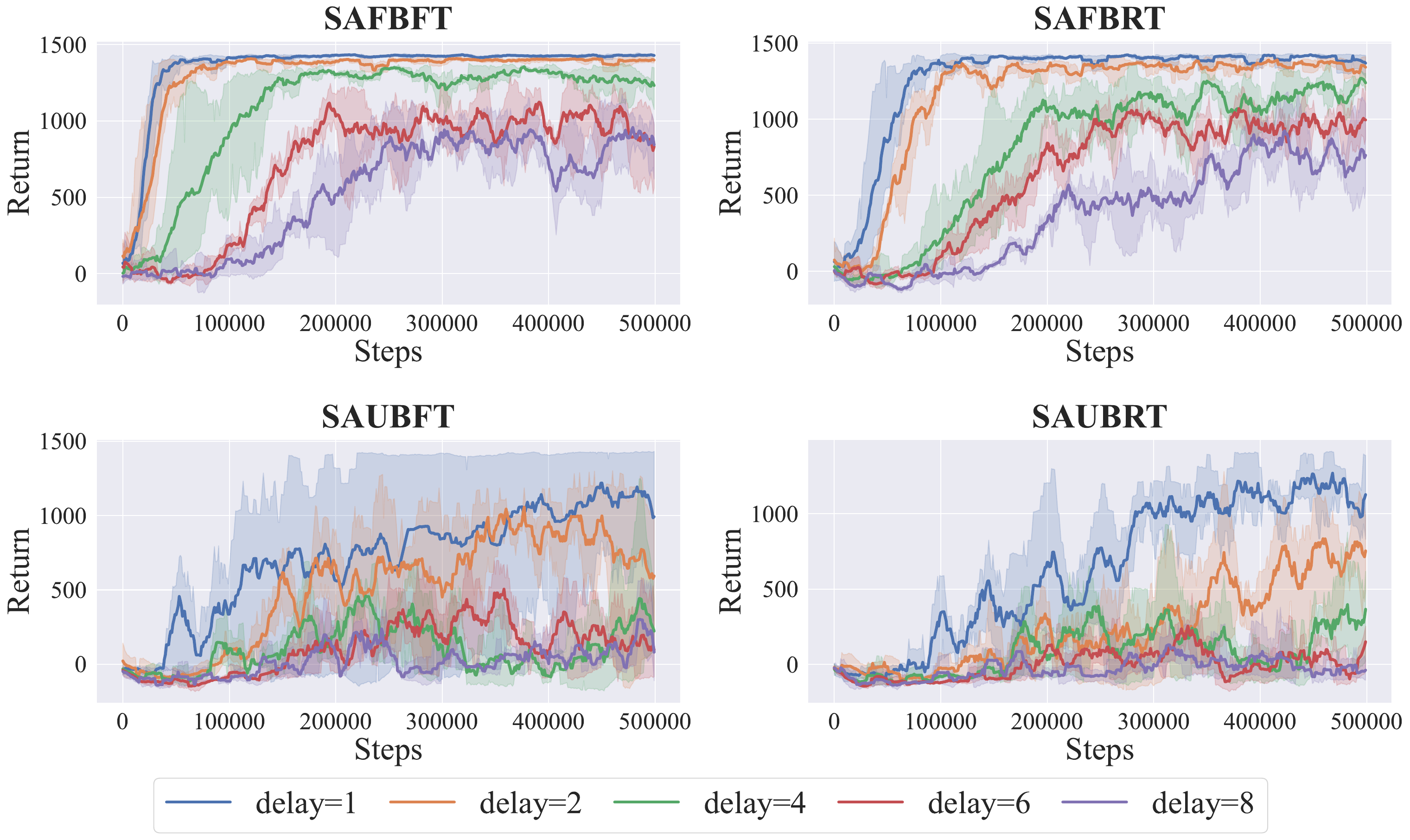 }
    \caption{\ Comparison of training with Prediction for different tasks under random delay.}
    \label{Appendix:rp}
\end{figure}

\begin{figure}[!h]
    \centering
    \includegraphics[width=0.95\textwidth]{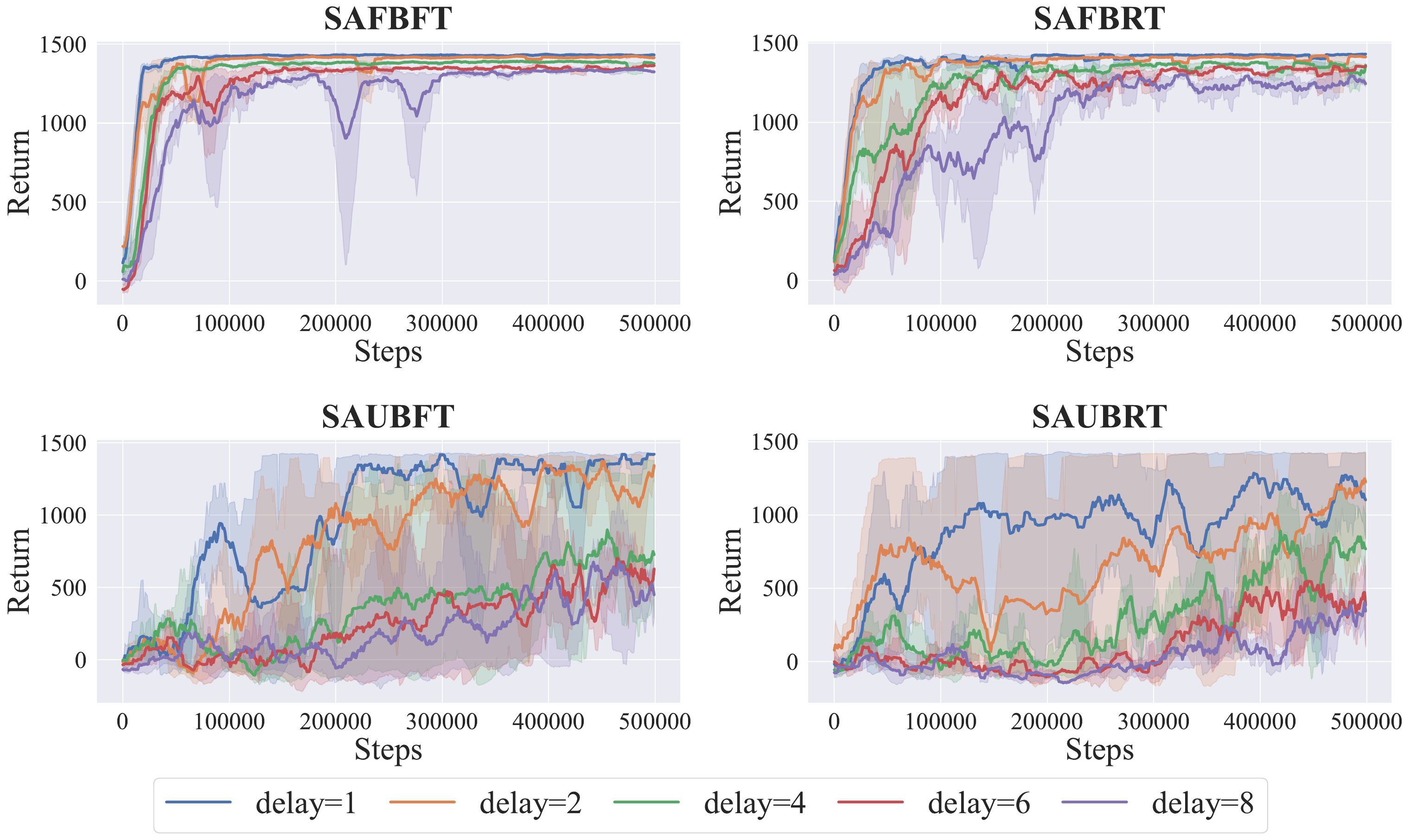 }
    \caption{\ Comparison of training with SACAS for different tasks under random delay.}
    \label{Appendix:rs}
\end{figure}



\newpage


\begin{thebibliography}{00}


\bibitem{gao2023hand}
Q.~Gao, J.~Li, Y.~Zhu, S.~Wang, J.~Liufu, J.~Liu, Hand gesture teleoperation for dexterous manipulators in space station by using monocular hand motion capture, Acta Astronautica 204 (2023) 630--639.

\bibitem{pryor2020interactive}
W.~Pryor, B.~P. Vagvolgyi, A.~Deguet, S.~Leonard, L.~L. Whitcomb, P.~Kazanzides, Interactive planning and supervised execution for high-risk, high-latency teleoperation, in: 2020 IEEE/RSJ International Conference on Intelligent Robots and Systems (IROS), IEEE, 2020, pp. 1857--1864.

\bibitem{zhang2022review}
W.~Zhang, F.~Li, J.~Li, Q.~Cheng, Review of on-orbit robotic arm active debris capture removal methods, Aerospace 10~(1) (2022) 13.

\bibitem{doggett2002robotic}
W.~Doggett, Robotic assembly of truss structures for space systems and future research plans, in: Proceedings, IEEE Aerospace Conference, Vol.~7, IEEE, 2002, pp. 7--7.

\bibitem{wang2008general}
X.~Wang, W.~Xu, B.~Liang, C.~Li, General scheme of teleoperation for space robot, in: 2008 IEEE/ASME International Conference on Advanced Intelligent Mechatronics, IEEE, 2008, pp. 341--346.

\bibitem{zhang2017autonomous}
X.~Zhang, J.~Liu, Autonomous trajectory planner for space telerobots capturing space debris under the teleprogramming framework, Advances in Mechanical Engineering 9~(9) (2017) 1687814017723298.

\bibitem{funda1992teleprogramming}
J.~Funda, T.~S. Lindsay, R.~P. Paul, Teleprogramming: Toward delay-invariant remote manipulation, Presence: Teleoperators \& Virtual Environments 1~(1) (1992) 29--44.

\bibitem{nuno2011passivity}
E.~Nu{\~n}o, L.~Basa{\~n}ez, R.~Ortega, Passivity-based control for bilateral teleoperation: A tutorial, Automatica 47~(3) (2011) 485--495.

\bibitem{wang2018novel}
Z.~Wang, Z.~Chen, B.~Liang, B.~Zhang, A novel adaptive finite time controller for bilateral teleoperation system, Acta Astronautica 144 (2018) 263--270.

\bibitem{zhai2016adaptive}
D.-H. Zhai, Y.~Xia, Adaptive finite-time control for nonlinear teleoperation systems with asymmetric time-varying delays, International Journal of Robust and Nonlinear Control 26~(12) (2016) 2586--2607.

\bibitem{chen2014adaptive}
Z.~Chen, Y.-J. Pan, J.~Gu, Adaptive robust control of bilateral teleoperation systems with unmeasurable environmental force and arbitrary time delays, IET Control Theory \& Applications 8~(15) (2014) 1456--1464.

\bibitem{chen2016self}
Z.~Chen, B.~Liang, T.~Zhang, A self-adjusting compliant bilateral control scheme for time-delay teleoperation in constrained environment, Acta Astronautica 122 (2016) 185--195.

\bibitem{mersha2013bilateral}
A.~Y. Mersha, S.~Stramigioli, R.~Carloni, On bilateral teleoperation of aerial robots, IEEE Transactions on Robotics 30~(1) (2013) 258--274.

\bibitem{sharifi2018impedance}
M.~Sharifi, H.~Salarieh, S.~Behzadipour, M.~Tavakoli, Impedance control of non-linear multi-dof teleoperation systems with time delay: absolute stability, IET Control Theory \& Applications 12~(12) (2018) 1722--1729.

\bibitem{kheddar2007enhanced}
A.~Kheddar, E.-S. Neo, R.~Tadakuma, K.~Yokoi, Enhanced teleoperation through virtual reality techniques, Advances in telerobotics (2007) 139--159.

\bibitem{deng2003predictive}
Z.~Deng, M.~Jagersand, Predictive display system for tele-manipulation using image-based modeling and rendering, in: Proceedings 2003 IEEE/RSJ International Conference on Intelligent Robots and Systems (IROS 2003)(Cat. No. 03CH37453), Vol.~3, IEEE, 2003, pp. 2797--2802.

\bibitem{liu2015dynamics}
X.~Liu, H.~Li, J.~Wang, G.~Cai, Dynamics analysis of flexible space robot with joint friction, Aerospace science and technology 47 (2015) 164--176.

\bibitem{mnih2013playing}
V.~Mnih, K.~Kavukcuoglu, D.~Silver, A.~Graves, I.~Antonoglou, D.~Wierstra, M.~Riedmiller, Playing atari with deep reinforcement learning, arXiv preprint arXiv:1312.5602 (2013).

\bibitem{vinyals2017starcraft}
O.~Vinyals, T.~Ewalds, S.~Bartunov, P.~Georgiev, A.~S. Vezhnevets, M.~Yeo, A.~Makhzani, H.~K{\"u}ttler, J.~Agapiou, J.~Schrittwieser, et~al., Starcraft ii: A new challenge for reinforcement learning, arXiv preprint arXiv:1708.04782 (2017).

\bibitem{park2022control}
J.~Park, T.~Kim, S.~Seong, S.~Koo, Control automation in the heat-up mode of a nuclear power plant using reinforcement learning, Progress in Nuclear Energy 145 (2022) 104107.

\bibitem{nian2020review}
R.~Nian, J.~Liu, B.~Huang, A review on reinforcement learning: Introduction and applications in industrial process control, Computers \& Chemical Engineering 139 (2020) 106886.

\bibitem{ouyang2022training}
L.~Ouyang, J.~Wu, X.~Jiang, D.~Almeida, C.~Wainwright, P.~Mishkin, C.~Zhang, S.~Agarwal, K.~Slama, A.~Ray, et~al., Training language models to follow instructions with human feedback, Advances in Neural Information Processing Systems 35 (2022) 27730--27744.

\bibitem{yan2018control}
C.~Yan, Q.~Zhang, Z.~Liu, X.~Wang, B.~Liang, Control of free-floating space robots to capture targets using soft q-learning, in: 2018 IEEE International Conference on Robotics and Biomimetics (ROBIO), IEEE, 2018, pp. 654--660.

\bibitem{wang2021end}
S.~Wang, Y.~Cao, X.~Zheng, T.~Zhang, An end-to-end trajectory planning strategy for free-floating space robots, in: 2021 40th Chinese Control Conference (CCC), IEEE, 2021, pp. 4236--4241.

\bibitem{lei2022active}
W.~Lei, H.~Fu, G.~Sun, Active object tracking of free floating space manipulators based on deep reinforcement learning, Advances in Space Research 70~(11) (2022) 3506--3519.

\bibitem{wu2020reinforcement}
Y.-H. Wu, Z.-C. Yu, C.-Y. Li, M.-J. He, B.~Hua, Z.-M. Chen, Reinforcement learning in dual-arm trajectory planning for a free-floating space robot, Aerospace Science and Technology 98 (2020) 105657.

\bibitem{wang2021multi}
S.~Wang, X.~Zheng, Y.~Cao, T.~Zhang, A multi-target trajectory planning of a 6-dof free-floating space robot via reinforcement learning, in: 2021 IEEE/RSJ International Conference on Intelligent Robots and Systems (IROS), IEEE, 2021, pp. 3724--3730.

\bibitem{cao2023reinforcement}
Y.~Cao, S.~Wang, X.~Zheng, W.~Ma, X.~Xie, L.~Liu, Reinforcement learning with prior policy guidance for motion planning of dual-arm free-floating space robot, Aerospace Science and Technology 136 (2023) 108098.

\bibitem{li2021constrained}
Y.~Li, X.~Hao, Y.~She, S.~Li, M.~Yu, Constrained motion planning of free-float dual-arm space manipulator via deep reinforcement learning, Aerospace Science and Technology 109 (2021) 106446.

\bibitem{wang2022collision}
S.~Wang, Y.~Cao, X.~Zheng, T.~Zhang, Collision-free trajectory planning for a 6-dof free-floating space robot via hierarchical decoupling optimization, IEEE Robotics and Automation Letters 7~(2) (2022) 4953--4960.

\bibitem{wang2022learning}
S.~Wang, Y.~Cao, X.~Zheng, T.~Zhang, A learning system for motion planning of free-float dual-arm space manipulator towards non-cooperative object, Aerospace Science and Technology 131 (2022) 107980.

\bibitem{yang2019control}
C.~Yang, J.~Yang, X.~Wang, B.~Liang, Control of space flexible manipulator using soft actor-critic and random network distillation, in: 2019 IEEE International Conference on Robotics and Biomimetics (ROBIO), IEEE, 2019, pp. 3019--3024.

\bibitem{jiang2021coordinated}
D.~Jiang, Z.~Cai, H.~Peng, Z.~Wu, Coordinated control based on reinforcement learning for dual-arm continuum manipulators in space capture missions, Journal of Aerospace Engineering 34~(6) (2021) 04021087.

\bibitem{katsikopoulos2003markov}
K.~V. Katsikopoulos, S.~E. Engelbrecht, Markov decision processes with delays and asynchronous cost collection, IEEE transactions on automatic control 48~(4) (2003) 568--574.

\bibitem{nath2021revisiting}
S.~Nath, M.~Baranwal, H.~Khadilkar, Revisiting state augmentation methods for reinforcement learning with stochastic delays, in: Proceedings of the 30th ACM International Conference on Information \& Knowledge Management, 2021, pp. 1346--1355.

\bibitem{bouteiller2021reinforcement}
Y.~Bouteiller, S.~Ramstedt, G.~Beltrame, C.~Pal, J.~Binas, Reinforcement learning with random delays, in: International conference on learning representations, 2021.

\bibitem{xie2023addressing}
M.~Xie, B.~Xia, Y.~Yu, X.~Wang, Y.~Chang, Addressing delays in reinforcement learning via delayed adversarial imitation learning, in: International Conference on Artificial Neural Networks, Springer, 2023, pp. 271--282.

\bibitem{liotet2022delayed}
P.~Liotet, D.~Maran, L.~Bisi, M.~Restelli, Delayed reinforcement learning by imitation, in: International Conference on Machine Learning, PMLR, 2022, pp. 13528--13556.

\bibitem{walsh2009learning}
T.~J. Walsh, A.~Nouri, L.~Li, M.~L. Littman, Learning and planning in environments with delayed feedback, Autonomous Agents and Multi-Agent Systems 18 (2009) 83--105.

\bibitem{hester2013texplore}
T.~Hester, P.~Stone, Texplore: real-time sample-efficient reinforcement learning for robots, Machine learning 90 (2013) 385--429.

\bibitem{firoiu2018human}
V.~Firoiu, T.~Ju, J.~Tenenbaum, At human speed: Deep reinforcement learning with action delay, arXiv preprint arXiv:1810.07286 (2018).

\bibitem{derman2020acting}
E.~Derman, G.~Dalal, S.~Mannor, Acting in delayed environments with non-stationary markov policies, in: International Conference on Learning Representations, 2020.

\bibitem{chen2021delay}
B.~Chen, M.~Xu, L.~Li, D.~Zhao, Delay-aware model-based reinforcement learning for continuous control, Neurocomputing 450 (2021) 119--128.

\bibitem{xia2024deer}
B.~Xia, Y.~Kong, Y.~Chang, B.~Yuan, Z.~Li, X.~Wang, B.~Liang, Deer: A delay-resilient framework for reinforcement learning with variable delays, arXiv preprint arXiv:2406.03102 (2024).

\bibitem{schuitema2010control}
E.~Schuitema, L.~Bu{\c{s}}oniu, R.~Babu{\v{s}}ka, P.~Jonker, Control delay in reinforcement learning for real-time dynamic systems: A memoryless approach, in: 2010 IEEE/RSJ International Conference on Intelligent Robots and Systems, IEEE, 2010, pp. 3226--3231.

\bibitem{agarwal2021blind}
M.~Agarwal, V.~Aggarwal, Blind decision making: Reinforcement learning with delayed observations, Pattern Recognition Letters 150 (2021) 176--182.

\bibitem{umetani1989resolved}
Y.~Umetani, K.~Yoshida, Resolved motion rate control of space manipulators with generalized jacobian matrix, Ph.D. thesis, Tohoku University (1989).

\bibitem{liu2021probabilistic}
Y.~Liu, X.~Wang, Z.~Tang, N.~Qi, Probabilistic ensemble neural network model for long-term dynamic behavior prediction of free-floating space manipulators, Aerospace Science and Technology 119 (2021) 107138.


\end{thebibliography}


\end{document}